\documentclass[twocolumn]{article}

\usepackage[utf8]{inputenc} % allow utf-8 input
\usepackage[T1]{fontenc}    % use 8-bit T1 fonts
\usepackage{hyperref}       % hyperlinks
\usepackage{url}            % simple URL typesetting
\usepackage{booktabs}       % professional-quality tables
\usepackage{amsfonts}       % blackboard math symbols
\usepackage{nicefrac}       % compact symbols for 1/2, etc.
\usepackage{microtype}      % microtypography
\usepackage{makecell}
\usepackage{amssymb}
\usepackage{bm}
\usepackage{adjustbox}
\usepackage{multirow}
\usepackage{mathtools}
\usepackage{subcaption}
\usepackage{lineno}
\usepackage{setspace}
\let\labelindent\relax   %To avoid error message with ieeecolor
\usepackage{enumitem}
\usepackage{amsmath}
\usepackage{comment}
\usepackage{bbm}
\usepackage[table]{xcolor}% http://ctan.org/pkg/xcolor
\newcommand\MyBox[1]{
  \fbox{\lower0.75cm
    \vbox to 1.7cm{\vfil
      \hbox to 1.7cm{\hfil\parbox{1.4cm}{\centering #1}\hfil}
      \vfil}%
  }%
}
\usepackage[T1]{fontenc}
\usepackage[table]{xcolor}
\usepackage{collcell}
\usepackage{hhline}
\usepackage{pgf}

\def\colorModel{hsb} %You can use rgb or hsb

\newcommand\ColCell[1]{
  \pgfmathparse{#1<100?1:0}  %Threshold for changing the font color into the cells
    \ifnum\pgfmathresult=0\relax\color{white}\fi
  \pgfmathsetmacro\compA{0}      %Component R or H
  \pgfmathsetmacro\compB{0}      %Component G or S
  \pgfmathsetmacro\compC{1 - #1/200}      %Component B or B
  \edef\x{\noexpand\centering\noexpand\cellcolor[\colorModel]{\compA,\compB,\compC}}\x #1 \%
  } 
\newcolumntype{E}{>{\collectcell\ColCell}m{2cm}<{\endcollectcell}}  %Cell width

\usepackage{authblk}

\usepackage{lscape}
\usepackage{arydshln} %dashed line

\newcommand{\N}{\mathcal{N}} %gaussian normal
\newcommand{\M}{\mathcal{M}} %Conjunto
\DeclarePairedDelimiter\p{(}{)}

\DeclarePairedDelimiter\llav{\{}{\}}

\DeclareMathOperator{\Xm}{\mathbf{X}^{(m)}} %X^(m)
\DeclareMathOperator{\XM}{\mathbf{X}^{\{\M\}}} %X^(M)
\DeclareMathOperator{\Xone}{\mathbf{X}^{(1)}} %X^(1)
\DeclareMathOperator{\Xtwo}{\mathbf{X}^{(2)}} %X^(2)
\DeclareMathOperator{\Xsix}{\mathbf{X}^{(6)}} %X^(1)
\DeclareMathOperator{\Xtwelve}{\mathbf{X}^{(12)}} %X^(1)
\DeclareMathOperator{\Xthirteen}{\mathbf{X}^{(13)}} %X^(1)
\DeclareMathOperator{\Xfourteen}{\mathbf{X}^{(14)}} %X^(1)
\DeclareMathOperator{\Xnm}{\mathbf{x}_{n,:}^{(m)}} %X_n^(m)
\DeclareMathOperator{\XnMM}{\mathbf{x}_{n,:}^{\{\M\}}} %X_n^(m)
\DeclareMathOperator{\XnM}{\mathbf{x}_{n,:}^{(M)}} %X_n^(m)
\DeclareMathOperator{\Xnone}{\mathbf{x}_{n,:}^{(1)}} %X_n^(1)
\DeclareMathOperator{\Z}{\mathbf{Z}} %Y
\DeclareMathOperator{\Zn}{\mathbf{z}_{n,:}} %Y
\DeclareMathOperator{\Wm}{\mathbf{W}^{(m)}} %W^(m)
\DeclareMathOperator{\Wone}{\mathbf{W}^{(1)}} %W^(1)
\DeclareMathOperator{\WM}{\mathbf{W}^{(M)}} %W^(1)
\DeclareMathOperator{\WmT}{\mathbf{W}^{(m)^T}} %W^(m)^T
\DeclareMathOperator{\WMM}{\mathbf{W}^{\{\M\}}} %W^(m)^T
\DeclareMathOperator{\Wkm}{\mathbf{w}_{:,k}^{(m)}} %\mathbf{w}_:k^(m)
\DeclareMathOperator{\Wdkm}{w_{d,k}^{(m)}} %w_dk^(m)
\DeclareMathOperator{\akm}{\alpha_k^{(m)}} %alpha_k^(m)
\DeclareMathOperator{\gamm}{\text{\boldmath$\gamma$}^{(m)}} %gamma^(m)
\DeclareMathOperator{\gamdm}{\gamma_d^{(m)}} %gamma_d^(m)
\DeclareMathOperator{\taum}{\tau^{(m)}} %tau^(m)
\DeclareMathOperator{\tauMM}{\tau^{\{\M\}}} %tau^(m)
\DeclareMathOperator{\tauone}{\tau^{(1)}} %tau^(1)
\DeclareMathOperator{\tauM}{\tau^{(M)}} %tau^(1)
\DeclareMathOperator{\sumn}{\sum\limits_{n=1}^{N}} %sum_{n=1}^{N}
\DeclareMathOperator{\prodd}{\prod\limits_{d=1}^{D_m}} %prod_{d=1}^{D_m}
\DeclareMathOperator{\prodn}{\prod\limits_{n=1}^{N}} %prod_{n=1}^{N}
\DeclareMathOperator{\prodk}{\prod\limits_{k=1}^{K_c}} %prod_{k=1}^{K_c}
\DeclareMathOperator{\prodm}{\prod\limits_{m=1}^{M}} %prod_{m=1}^{2}

\newcommand{\eqline}{\enskip&\enskip}

\newcommand{\eqsimil}{\enskip\sim &\enskip}

\usepackage{ulem}
\usepackage{xcolor}
\newcommand{\carlos}[1]{{\color{black}{#1}}}

\usepackage{cite}
\usepackage{algorithmic}
\usepackage{graphicx}
\usepackage{textcomp}
\usepackage{tabularx} % To adjust to column width

\makeatletter
\def\@fnsymbol#1{\ensuremath{\ifcase#1\or \dagger\or \ddagger\or
   \mathsection\or \mathparagraph\or \|\or **\or \dagger\dagger
   \or \ddagger\ddagger \else\@ctrerr\fi}}
   
\begin{document}
\title{Multi-task longitudinal forecasting with missing values on Alzheimer's Disease\thanks{Data used in preparation of this article were obtained from the Alzheimer’s Disease Neuroimaging Initiative (ADNI) database (adni.loni.usc.edu). As such, the investigators within the ADNI contributed to the design and implementation of ADNI and/or provided data but did not participate in analysis or writing of this report. This work involved human subjects or animals in its research. Approval of all ethical and experimental procedures and protocols was granted by the Institutional Data Access/Ethics Committee of the ADNI. A complete listing of ADNI investigators can be found at: \url{http://adni.loni.usc.edu/wpcontent/uploads/how_to_apply/ADNI_Acknowledgement_List.pdf}. This work involved human subjects or animals in its research. Approval of all ethical and experimental procedures and protocols was granted by the Institutional Data Access/Ethics Committee of the Alzheimer’s Disease Neuroimaging Initiative (ADNI).}}
\author[1,*]{Carlos Sevilla-Salcedo}
\author[2]{Vandad Imani}
\author[1]{Pablo M. Olmos}
\author[1]{Vanessa G\'omez-Verdejo}
\author[2]{Jussi Tohka }
\affil[1]{Department of Signal Processing and Communications, Universidad Carlos III de Madrid, Legan\'es, 28911 Spain}
\affil[2]{A.I. Virtanen Institute for Molecular Sciences, University of Eastern Finland, Kuopio, Finland}
\affil[*]{Corresponding author: Carlos Sevilla-Salcedo, sevisal@tsc.uc3m.es}
% }
\maketitle

\begin{abstract}
Machine learning techniques typically applied to dementia forecasting lack in their capabilities to jointly learn several tasks, handle time dependent heterogeneous data and missing values. In this paper, we propose a framework using the recently presented SSHIBA model for jointly learning different tasks on longitudinal data with missing values. The method uses Bayesian variational inference to impute missing values
% , model both real-valued and categorical data 
and combine information of several views. This way, we can combine different data-views from different time-points in a common latent space and learn the relations between each time-point while simultaneously modelling and predicting several output variables. We apply this model to predict together diagnosis, ventricle volume, and clinical scores in dementia. The results demonstrate that SSHIBA is capable of learning a good imputation of the missing values and outperforming the baselines while simultaneously predicting three different tasks.
 
\end{abstract}


\setcounter{footnote}{0}

\section{Introduction}

% AD context
Alzheimer's Disease (AD) is a common form of dementia that manifests in the form of cognitive degeneration and conduct disorder. Although its symptoms vary between subjects, it is commonly characterised by memory loss as well as general cognitive decline.  More than 30 million people suffer from AD currently and this number is expected to triple by 2050\cite{barnes2011projected}. The number of people affected by the disease is higher than the number of AD patients due to the huge impact on the lives of relatives, friends and care-givers. AD has no cure, but interventions taking place early on during the disease cascade can improve the quality of life and alleviate the symptoms \cite{lisko2021can}. For this reason, the investigations for an early detection of AD in high risk individuals are critical. Similarly, cognitive scores are essential for understanding the efficacy of antidementia treatments as well as the disease progression\cite{kueper2018alzheimer}.  Machine Learning (ML) techniques can be used for the design of imaging biomarkers for various brain disorders and, additionally, the inferred ML models can be analysed as multivariate, discriminative representations of the brain disease.

% Multitask in AD

%Wang et al.\cite{wang2020modeling} formulated the progression of AD as a weakly supervised temporal multitask matrix regression framework that considers the prediction of cognitive scores at each time point as a regression task. 

%Various studies have formulated the prediction of cognitive decline in AD as a MultiTask Learning (MTL) problem. These studies generally fall into one of two data prediction problems: (1) multitask regression\cite{} and (2) multitask classification\cite{yang2019fused}. The main goal of MTL is to identify the essential relation across tasks and build learning models capable of capturing this information. MTL methods with sparsity-inducing norm regularisation have been widely studied to improve generalisation performance by simultaneously solving multiple learning tasks with shared parameters across tasks, e.g. $\ell_{2,1}$-norm regularisation. 
% MTL methods with sparsity-inducing norm regularization have been widely studied to improve generalization performance by simultaneously solving multiple learning tasks while utilizing commonalities across tasks. For example, the $\ell_{2,1}$-norm regularization enforces multiple predictors to have common sparsity patterns across all tasks. 
% Cao et al.~\cite{cao20182} proposed a $\ell_{2,1}$ -$\ell_{1}$-norm regularized multi-kernel MTL feature learning formulation with a joint sparsity inducing regularization to look for the common representation that are useful for modeling the disease’s cognitive scores. 

% Longitudinal information in AD
Analysis of the progression of dementia in longitudinal studies has been proven to be critical for an adequate treatment\cite{sevigny2016antibody}. Some algorithms primarily use neuroimaging information to analyse the disease progression of the disease\cite{duara2008medial,burton2009medial}. Koval et al.\cite{koval2018spatiotemporal} used graph nodes to represent a spatially structured mixed-effect model with a Monte-Carlo Markov-Chain Stochastic Approximation Expectation-Maximization to model the evolution of longitudinal brain imaging data. In contrast, other studies focus on analysing the progression of biomarkers to characterise the disease. Venkatraghavan  et al.\cite{venkatraghavan2021progression} construct a timeline of biomarker changes and models the brain abnormality with APOE genotypes using Gaussian Mixture Models to then calculate the probability of abnormality. Similarly, Donohue et al.\cite{donohue2014estimating} assume that the biomarkers are a set of curves with a common shape combined with simple linear effects at the subject level, while modelling long-term traits with nonparametric monotonic smoothing.  However, some studies combine both neuroimaging and biomarker information in a single framework to model the longitudinal effect of AD. Fonteijn et al.\cite{fonteijn2012event} combines heterogeneous data using both imaging and clinical data by defining the disease as a sequence of discrete events using a Bayesian model.

% Imputation specially with longitudinal data
Longitudinal dementia studies faces issues of missing data due to old age and health related concerns in the studied population\cite{hardy2009missing}. This problem is greatly accentuated when working with longitudinal data, where follow up measures are often interrupted either by cognitive impairment or mortality\cite{atkinson2007cognitive}. To work around the issue, some studies remove all the data from subjects with missing values. However, this reduces the number of usable data samples and can introduce bias in the data\cite{marti2020survey}. For this reason, other studies\cite{zhou2013modeling} apply diverse techniques to impute the missing data. Some use basic inference techniques such as substituting the missing variable by its mean, median or mode value. Others exploit the longitudinal nature of the problem to infer the missing values using temporal imputation, using available information at previous months\cite{huang2017longitudinal}. Adhikari et al.\cite{adhikari2019high} combines both imputation techniques, using temporal inference if previous data is available and the variable median otherwise. Bayesian algorithms assume variables are random and learn their distribution, which can be used to impute these missing values, for instance, using the mean of the distribution or sampling from the distribution\cite{bartolucci2009examination,mccombe2021practical}.

% Multitask in AD
Multitask learning (MTL) is a sub-field of ML that simultaneously learns multiple tasks by jointly optimising multiple loss functions. %MTL models improve generalization by combining auxiliary information from the different tasks. 
MTL models improve generalisation by leveraging the information contained in the training data of related tasks. It is, therefore, beneficial when the tasks have some level of correlation. In recent years, MTL has attracted a lot of attention in the prediction of the progression of cognitive decline in dementia at multiple time points with clinical data \cite{zhou2012modeling,zhou2013modeling,jie2015manifold,emrani2017prognosis,imani2021comparison}. The fundamental hypothesis in the longitudinal analysis is that the subject's clinical data cannot be assumed to be  independent at consecutive visits. Accordingly, the MTL can benefit the prediction of disease progression by capturing relatedness and shared information between several observation records across the visits. One of the critical issues in MTL is identifying the inherent relation between these records of observation and tasks. MTL methods with sparsity-inducing norm regularisation have been widely studied to improve generalisation performance by simultaneously solving multiple learning tasks while utilising commonalities across tasks. For example, Zhou et al.~\cite{zhou2013modeling} developed fused group Lasso formulation as the regularisation term to capture intrinsic temporal relationship among tasks from various time points to model the progression of AD. Lei et al.~\cite{lei2017longitudinal} employed joint sparsity regularisation ($\ell_{2,1}$-norm) in order to exploit a common subset of features for multiple longitudinal predictions of AD progression. Cao et al.~\cite{cao20182} proposed a $\ell_{2,1}$ -$\ell_{1}$-norm regularised multi-kernel MTL feature learning formulation with a joint sparsity inducing regularisation (SMKMTL). Their framework uses a mixed sparsity-Inducing $\ell_{2,1}$ -$\ell_{1}$-norm to capture the inherent correlation among the tasks. They have proposed SMKMTL multitask learning method to capture the kernel-wise association between MRI neuroimaging features and cognitive scores. Yang et al.~\cite{yang2019fused} proposed a fused sparse network algorithm with parameter-free centralised learning to model and identify the longitudinal analysis of early MCI and late MCI based on resting-state functional MRI. Tabarestani et al.~\cite{tabarestani2020distributed} proposed a multitask multimodal framework for predicting cognitive measures in the progression of AD. They applied $\ell$1-norm regularisation to introduce sparsity among all features and capture different modalities' inherent temporal sparsity patterns and their relative correlation strength.

However, we are not aware of studies or methods that analyse cognitive decline and combine MTL, longitudinal data and missing data imputation within a single framework. To address this, we propose developing a model based on the recently presented SSHIBA framework \cite{sevilla2021sparse} to predict various facets of cognitive decline. This model has the ability to work with multiple views and impute missing values. Specifically, we propose establishing different time-points as different views in order to model temporal relations and make a forecast for future time-points. This allows the model to find a common latent representation of time-dependent and time-independent variables that describes the temporal relation between variables over time. \carlos{Furthermore, the multi-view formulation of the framework allows having regression MTL.}
% Furthermore, the heterogeneous formulation of the framework allows to model data from different natures, having a simultaneous MTL of regression and classification prediction tasks. 
% Besides, the multi-view definition of the model as well as it semi-supervised formulation endows the model with multi-task learning, allowing to combine both regression and classification tasks. 
% Combining regression and classification tasks has received only a limited amount of attention in the literature, and the few models that do it define independent models for the regression and classification tasks\cite{zhang2012multi}. 
SSHIBA eliminates the need to train a model for each task and enhances the performance by combining the information of the multiple tasks in the common framework.
% Treating missing values in test datasets is mostly not addressed in the literature, even though it is possible to have missing values\cite{saar2007handling}. 
Finally, SSHIBA allows to impute all missing values in any view using its semi-supervised formulation to learn the joint variable distribution of the train and test datasets. This allows to automatically impute missing values using the learnt distribution and, simultaneously, using a semi-supervised scheme to predict the test data.

% In this paper, we propose using a recently presented model capable of combining multitask regression and classification with modelling longitudinal data and imputing missing values. SSHIBA, presented in Sevilla et al.\cite{sevilla2021sparse}, is a Bayesian model that finds latent data representations of heterogeneous data combined in different views. This way, the model can combine labelled and real data in different views and learn the temporal relations between them through their common representation in a low dimensional space. Treating missing values in test datasets is mostly not addressed in the literature, even though it is possible to have missing values\cite{saar2007handling}. However, SSHIBA allows to impute all missing values using its semi-supervised formulation to learn the joint variable distribution of the train and test datasets.

To analyse the performance of SSHIBA, we use the Alzheimer's Disease Neuroimaging Initiative (ADNI) database. Based on previous research, we select a subset of relevant variables for characterising AD to analyse the time evolution of the subject's diagnosis, ADAS-cog13 score (ADAS13), and ventricle volume. We measure the performance against several baseline methods for the imputation of missing values and the multitask prediction of the output variables.
The results indicate that SSHIBA outperforms the baselines on the prediction task while finding a latent representation of the data that improves the interpretability of the predictions.

\section{Materials}

Data used in the preparation of this article were obtained from the Alzheimer's Disease Neuroimaging Initiative (ADNI) database (\url{adni.loni.usc.edu}). The ADNI project was launched in 2003 as a public-private partnership, led by Principal Investigator Michael W. Weiner, MD. The primary goal of ADNI has been to test whether serial magnetic resonance imaging (MRI), positron emission tomography (PET), other biological markers, and clinical and neuropsychological assessment can be combined to measure the longitudinal progression of Mild Cognitive Impairment (MCI) and early AD. In particular, we use the tables prepared for the TADPOLE grand challenge based on ADNI data (\url{https://tadpole.grand-challenge}) \cite{marinescu2018tadpole}. Although in the original database there is information for up to 120 months after baseline, for this article, we use the data corresponding to the first 36 months.

\begin{figure}[thb]
  \centering
    \includegraphics[width=\linewidth]{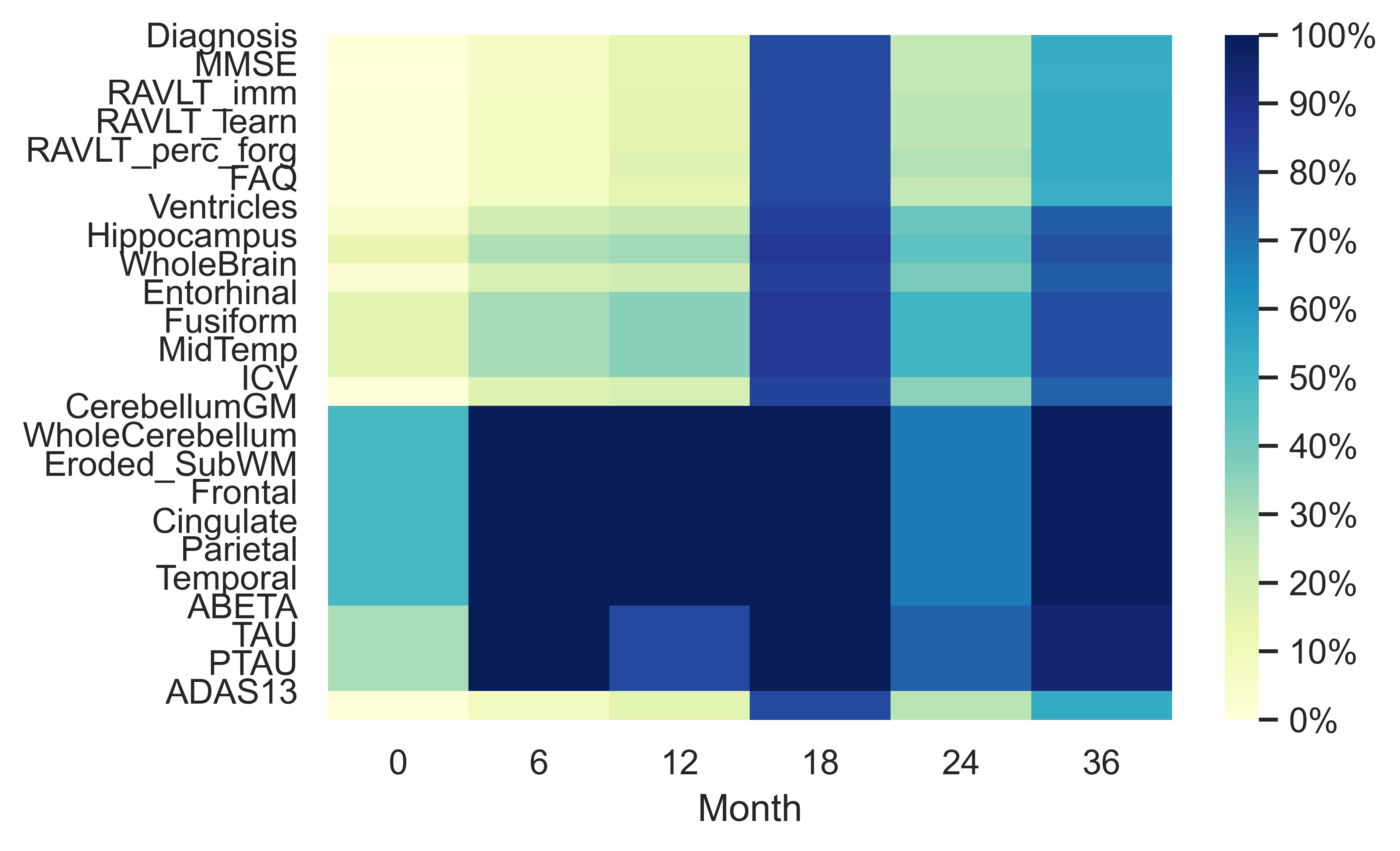}\\
  \caption{Heat map of the percentage of missing values for each TD variable for each month. Dark blue colours represent the variable is missing for all individuals at the specified month, while light yellow colours represent there are no missing values.}
  \label{fig:HeatMap}
\end{figure}

From the original database with $1,739$ subjects we selected the $1,730$ subjects with at least some information on month 36 described in Table \ref{tab:demographic}. Following \cite{prakash2020quantitative}, we use certain features that are specially relevant for the characterisation of the disease (see Table \ref{tab:Variables} for a summary of these features).  Figure \ref{fig:HeatMap} depicts the number of missing values for each Time Dependent (TD) variable at each analysed month. Note that the proportion of missing values is large for most features. This is because, although we use all variables at each visit (timepoint), not all variables are acquired at every visit and not all participants are scheduled visit at every 6 months. although we use all variables at each visit (timepoint), not all variables are acquired at every visit and not all participants are scheduled visit at every 6 months.

\begin{table}[h!t]
\arrayrulecolor{black}
\renewcommand{\arraystretch}{1.0}
\caption{Demographic information of participants. The diagnosis is the diagnosis at the baseline. The column age lists the average age followed by the standard deviation of age.}
\label{tab:demographic}
\centering
\begin{adjustbox}{max width=\columnwidth}
\begin{tabular}{c c c c}
\hline
 Diagnosis & No. of Subjects  & Age & Male / Female  \\ 
\hline
NC & 523 & $74.22 \pm 5.80$ & 253 / 270 \\
MCI & 866 & $73.05 \pm 7.60$ & 512 / 354 \\
AD & 341 & $74.98 \pm 7.76$ & 189 / 152 \\
\hline
\end{tabular}
\end{adjustbox}
\end{table}

\begin{table}[h!t]
\arrayrulecolor{black} %Table line colors
\renewcommand{\arraystretch}{1.0}
\caption{Description of the variables used in this study. Each variable is assigned to one group to facilitate the understanding of the framework: \textit{TI}, time-independent variables; \textit{TD}, time-dependent variables; \textit{D}, Diagnosis; \textit{V}, Ventricle volume; \textit{A}, ADAS13 score. NePB indicates neuropsychology and behavioral tests and we use FDG-PET at baseline. 
% Last column indicates the total percentage of missing values considering the first 36 months after baseline.
}
\label{tab:Variables}
\centering
\begin{adjustbox}{max width=\columnwidth}
\begin{tabular}{ccccc}
\toprule
Variable & Description & Group \\%& Miss.\\
 \midrule
% Age & Age at baseline & No & TI \\
% Education & Years of education & No & TI \\
% APOE4 & Number of APOE4 alleles & No & TI \\
% Gender & Female or Male & No & TI \\
Age                     & \multirow{4}{*}{Subject details}  & TI     \\%& $0.1\%$\\
Sex                     &                                   & TI     \\%& $0.1\%$\\
APOE4                   &                                   & TI     \\%& $0.2\%$\\
Education               &                                   & TI     \\%& $0.1\%$\\
\hline
AngularLeft             & \multirow{5}{*}{FDG-PET}          & TI     \\%& $5\%$\\
AngularRight            &                                   & TI     \\%& $5\%$\\
CingulumPostBilateral   &                                   & TI     \\%& $5\%$\\
TemporalLeft            &                                   & TI     \\%& $5\%$\\
TemporalRight           &                                   & TI     \\%& $5\%$\\
\hline
MMSE                    & \multirow{5}{*}{NePB}             & TD \\%& $50\%$\\
RAVLT learning          &                                   & TD \\%& $50\%$\\
RAVLT immediate         &                                   & TD \\%& $50\%$\\
RAVLT perc forgetting   &                                   & TD \\%& $50\%$\\
FAQ                     &                                   & TD \\%& $50\%$\\
\hline
Cerebellum Grey Matter  & \multirow{7}{*}{AVF45 data}       & TD \\%& $89\%$\\
Whole Cerebellum        &                                   & TD \\%& $89\%$\\
Eroded Subcortical Wm   &                                   & TD \\%& $89\%$\\
Frontal                 &                                   & TD \\%& $89\%$\\
Cingulate               &                                   & TD \\%& $89\%$\\
Parietal                &                                   & TD \\%& $89\%$\\
Temporal                &                                   & TD \\%& $89\%$\\
\hline
ABETA                   & \multirow{3}{*}{CSF values}       & TD \\%& $84\%$\\
TAU                     &                                   & TD \\%& $84\%$\\
PTAU                    &                                   & TD \\%& $84\%$\\
\hline
Hippocampus             & \multirow{6}{*}{MRI volumetry}    & TD \\%& $60\%$\\
WholeBrain              &                                   & TD \\%& $54\%$\\
Entorhinal              &                                   & TD \\%& $61\%$\\
Fusiform                &                                   & TD \\%& $61\%$\\
MidTemp                 &                                   & TD \\%& $61\%$\\
ICV                     &                                   & TD \\%& $53\%$\\
\hline
Ventricle volume        & MRI volumetry                     & V \\%& $55\%$\\
ADAS13                  & ADAS-Cog13 score                  & A \\%& $50\%$\\
Diagnosis               & Clinical diagnosis                & D \\%& $53\%$\\
\bottomrule
\end{tabular}
\end{adjustbox}
\end{table}

We considered various participant details that have been found to be risk factors contributing to AD\cite{duara1996alzheimer}: age, sex, the number of APOE e4 alleles, and the years of education. 
% With ageing, normal cognitive decline is an accepted phenomenon, but lower education and lower cerebral metabolic activity could accelerate the normal decline\cite{prencipe1996prevalence}. \carlos{Sex effect might be small to moderate, however\cite{ferretti2018sex} advises to consider it when designing predictive models for diagnosis of AD.} The APOE e4 allele, present in approximately 10-15$\%$ of people, increases the risk for late-onset AD and lowers the age of onset. One copy of e4 (e3/e4) can increase risk by 2-3 times while homozygotes (e4/e4) can be at 12 times increased risk\cite{michaelson2016apoe4}.
We coded APOE e4 status as either absence (0), single copy (1) or homozygous (2).

As FDG-PET features, we used average standardised uptake values (SUVs) in five brain regions \carlos{from the ADNI database}: bilateral angular gyri, bilateral posterior cingulate gyri, and bilateral inferior temporal gyri. The FDG-PET data measures glucose consumption and is shown to be strongly related to dementia and cognitive impairment when compared to normal control subjects\cite{landau2013comparing,landau2011associations,jagust2010alzheimer}. Motion correction and co-registration with MRI was performed on the acquired PET data. The highest 50$\%$ of voxel values within a hand-drawn pons/cerebellar vermis region were selected and their mean was used to normalise each ROI measurement resulting in the final FDG-PET measurements.
As specified in \cite{ortner2019amyloid, samuraki2007partial}, FDG-PET has been criticized as longitudinal biomarker for the analysis of cognitive decline, whereas AV-45 PET and MRI are more powerful in the longitudinal analysis of the disease. For this reason, we include only the FDG-PET values at baseline for these experiments.

The neuropsychology and behavioral (NePB) assessments reﬂect the cognitive abilities of the subjects. Subjects underwent a battery of NePB tests\cite{battista2017optimizing}. We included 5 NePB scores as features: the summary score from Mini-Mental State Examination (MMSE)\cite{folstein1975practical}, three summary scores of Rey’s auditory verbal learning test (RAVLT; learning, immediate, and percent forgetting)\cite{rey1958examen}, and a summary score from the functional activities questionnaire (FAQ)\cite{pfeffer1982measurement}.

As AV-45 PET features, we used SUVs in seven regions of interest (ROIs): frontal cortex, cingulate, lateral parietal cortex, lateral temporal cortex, cerebellum grey matter, whole cerebellum, and eroded subcortical white matter. The AV-45 PET measures amyloid-beta load in the brain. AV-45 PET imaging and preprocessing details are available at \url{http://adni.loni.usc.edu/methods/pet-analysis-method/pet-analysis/}. We used regional SUV ratios processed according to the UC Berkeley protocol\cite{johnson2013florbetapir,landau2012amyloid,landau2013comparing}. Each AV-45 PET scan was co-registered to the corresponding MRI and the mean AV-45 uptake within the regions of interest and reference regions was calculated.
% Regions of interest were composites of frontal regions, anterior/posterior cingulate regions, lateral parietal regions, and lateral temporal regions\cite{mormino2009episodic}. 
% The ﬁnal PET measurements were the average amyloid-beta uptakes in the four ROIs normalized by the whole cerebellum reference region.
We included the values in ROIs as well as values in the reference regions as variables to the model to learn to normalise the values as it remains uncertain which reference region would be the best for standardisation \cite{shokouhi2016reference}.
%There is discussion about what is the best method to normalise these images \cite{shokouhi2016reference}. Our approach is using the ML model to learn the normalisation.
% Claim SSHIBA normalises ROIs, (ICV) normalisation factors in ML model. Really new imaging technique, more degrees of freedom in the normalisation. 
% Discussion about the best method to normalise these images\cite{shokouhi2016reference}, ML can also learn the normalisation

The baseline Cerebrospinal Fluid (CSF) A$\beta$42, t-tau, and p-tau were used as CSF features\cite{shaw2009cerebrospinal}. 
% CSF was collected in the morning after an overnight fast using a 20- or 24-gauge spinal needle, frozen within 1 hour of collection, and transported on dry ice to the ADNI Biomarker Core laboratory at the University of Pennsylvania Medical Center. 
% The levels of A$\beta$42, t-tau, and p-tau in CSF were used.

As MRI features, we used 7 features: intracranial volume (ICV), and volumes of the hippocampus, entorhinal cortex, fusiform gyri, whole brain, middle temporal gyri and lateral ventricles. These features were selected based on previous studies\cite{gomez2018comparison}.
% We included volumes divided by the ICV as it is unclear whether raw or ICV-corrected volumes are better predictors of dementia\cite{gomez2018comparison,voevodskaya2014effects}. 
MRI protocol details are provided by ADNI at \url{http://adni.loni.usc.edu/methods/mri-tool/mri-analysis/}. Cortical reconstruction and volumetric segmentation had been performed with the FreeSurfer 5.1 image analysis suite.

The ADAS-cog 11 task scale was developed to assess the efficacy of anti-dementia treatments. Further developments to the scale shifted its sensitivity towards pre-dementia syndromes as well, primarily mild cognitive impairment (MCI). The ADAS-cog 13 task scale was one such improvement on the original ADAS-cog 11, with additional memory and attention/executive function tasks\cite{mohs1997development}. The 13 tasks test verbal memory (3 tasks), clinician-rated perception (4 tasks), and general cognition (6 tasks). It was found to perform better than the ADAS-cog 11 at discriminating between MCI and mild AD subjects, as well as have better sensitivity to treatment effects in MCI\cite{raghavan2013adas}. 
% As the ADAS-cog 13 fully encompasses the ADAS-cog 11 tasks, it is also backward compatible. 
As such, we used the ADAS-cog 13 scale for our study as a continuous quantitative measure of a subject’s disease status. 
% The scores at baseline, 12-month, 24-month and 36-month timelines were obtained from the ADNI dataset (Table S.2).
The value of these scores is lowest for the normal control group and increases with disease progression, with the highest scores for AD subjects.

As in the TADPOLE competition \cite{marinescu2018tadpole, marinescu2020alzheimer}, we consider: the clinical diagnosis (NC, MCI, AD) denoted as $D$, the ventricle volume of the MRI data denoted as $V$ and the ADAS-cog 13 score denoted as $A$. 

%%%%%%%%%%%%%%%%%%%%%%%%%%%%%%%%%%%%%%%%%%%%%%%%%%%%%%%%%%%%%%%%%%%%%%%%%%%%%%%%%%%%%%%%%%%%%%%%%%%%%%%%%%%%%%%%%%%%%%%%%%
%                                            Methodology 
%%%%%%%%%%%%%%%%%%%%%%%%%%%%%%%%%%%%%%%%%%%%%%%%%%%%%%%%%%%%%%%%%%%%%%%%%%%%%%%%%%%%%%%%%%%%%%%%%%%%%%%%%%%%%%%%%%%%%%%%%%

\section{Methods}
% *** I WOULD TERM SSHIBA AS "FRAMEWORK" AND RESULTING MODEL IN THIS PAPER AS "MODEL". THIS WOULD BE CLEARER AND I THINK THAT IT WOULD MAKE THE PAPER'S NOVELTY MORE APPARENT ***
% *** "REAL" SHOULD BE "REAL VALUED" 
% *** AVOID "BESIDES" -> MOST OFTEN IT SHOULD BE "IN ADDITION" , "ADDITIONALLY" , "FURTHERMORE" ETC.

In this work, we adapt the recently presented SSHIBA framework\cite{sevilla2021sparse} to model longitudinal data. SSHIBA is a Bayesian model able to combine multiview heterogeneous information into a common latent space. Here, we propose exploiting this multiview formulation to model the progression of the variables over time. 

\subsection{Review of SSHIBA}
\begin{figure}[t]
  \centering
%   \begin{subfigure}[t]{0.45\linewidth}
%     \centering
%     \includegraphics[width=\linewidth]{Images/Graphic_Model_Total.pdf}\\
%     % \includegraphics[page=1,width=0.6\textwidth]{images/Graphic_Model_Total.pdf}\\
%     \caption{Multi-view model.}
%     \label{fig:Schemea}
%   \end{subfigure}
%   ~
%   \begin{subfigure}[t]{0.41\linewidth}
    \centering
    \includegraphics[width=0.7\linewidth]{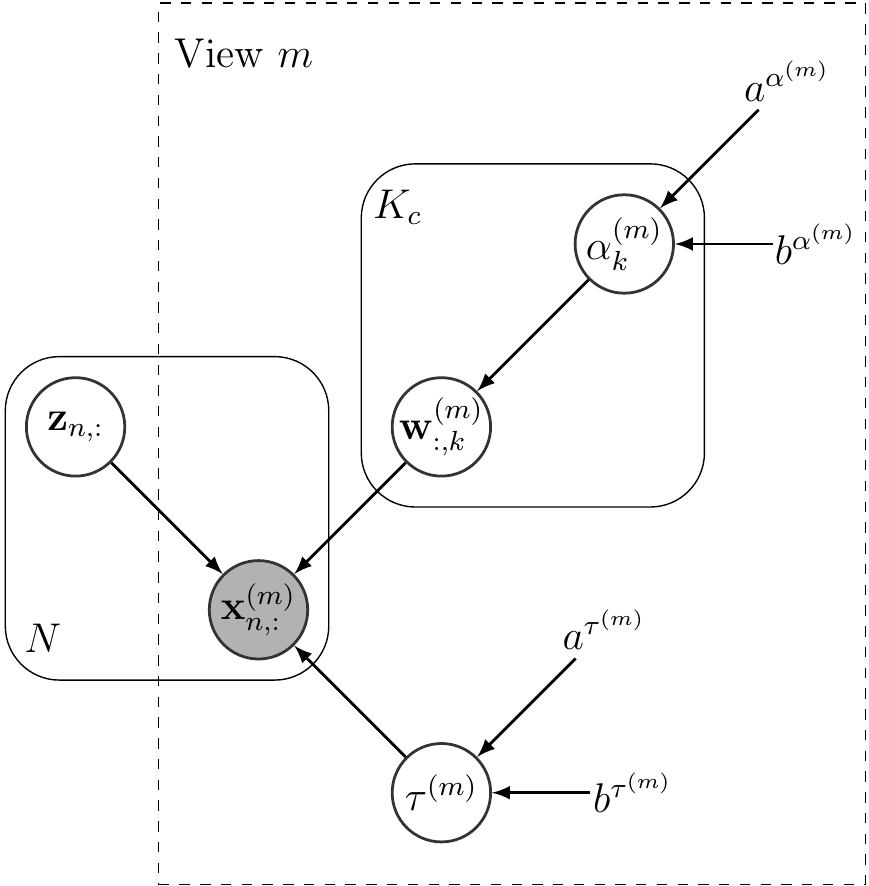}
%   \end{subfigure}
  \caption{Plate diagram for the SSHIBA graphical model with one view and real valued observations. Gray circles denote observed variables, white circles unobserved random variables. The nodes without a circle correspond to the hyperparameters.}
  \label{fig:Scheme}
\end{figure}

We consider a multi-view problem where we have $N$ data samples represented in $M$ different views, $\llav{\Xm}^{M}_{m=1}$. Therefore, we have that $\Xnm\in\mathbb{R}^{1 \times D_m}$ is the $m$-th view of the $n$-th subject with $n= 1,\ldots, N$ 
% (each view is a $D_m$-dimensional row vector)
and $\mathcal{M}=\{1,\ldots, M\}$, so $\XnMM = \{\Xnone,\ldots,\XnM\}$ is the complete $n$-th observation\footnote{Each view comprises an observation and the set of random variables associated in the model.}.
% , where $N$ is the number of subjects and $M$ is the number of views per subject
The model considers there is a latent variable, $\Z$, that can be combined with a set of projection matrices for each view, $\WMM = \{\Wone,\ldots,\WM\}$, and add some gaussian noise, with precision $\tauMM = \{\tauone,\ldots,\tauM\}$, to generate the complete observation $\XnMM$. Then, the probability density function (pdf) of the random variables for each view can be defined as
% of SSHIBA for the set of views $\mathcal{M}_o$ can be defined as:
\begin{align}
    \Zn \eqsimil \N(0,I_{K_c}) \label{eq:Zprior}\\
	\Wkm \eqsimil \N\p*{0,\p*{\akm}^{-1} I_{K_c}} \label{eq:Wprior}\\
    \Xnm | \Zn \eqsimil \N(\Zn \WmT, \taum^{-1} I_{D_m}) \label{eq:Xprior}\\
    \alpha_{k}^{(m)} \eqsimil \Gamma\p*{a^{\alpha^{(m)}}, b^{\alpha^{(m)}}} \label{eq:Alphaprior}\\
	\taum \eqsimil \Gamma\p*{ a^{\taum}, b^{\taum} } \label{eq:Tauprior}
\end{align}
where $I_{K_c}$ is an identity matrix of dimension $K_c$, $\Zn\in\mathbb{R}^{1 \times K_c}$ is the low-dimension latent variable for the $n$-th data point,  $\Gamma(a,b)$ is a Gamma distribution with parameters $a$ and $b$, $\Wkm$ is the $k$-th column of matrix $\Wm$ (of dimensions $D_m\times K_c$), and up-script $(m)$ corresponds to the $m$-th view. The Gamma distribution over $\alpha_{k}^{(m)}$ enables the model to enforce zero values in order to maximise the model likelihood given our data. Hence, we say that \eqref{eq:Wprior} and \eqref{eq:Alphaprior} form an ARD (Automatic Relevance Determination) prior\cite{neal2012Bayesian} for each column of matrix $\Wm$. The SSHIBA graphical model for the generation of each data view is included in Figure \ref{fig:Scheme}. 

These equations correspond to the standard formulation of SSHIBA to model real observations. However, if besides modelling real observations, we require using feature selection in a certain view, we can modify Equation \eqref{eq:Wprior} and add a new variable $\gamm$
\begin{align}
    \Wdkm  \eqsimil \N\p*{0,\p*{\gamdm\akm}^{-1}} \label{eq:Wprior_sparse} \\
    \gamdm \eqsimil \Gamma\p*{a^{\gamm}, b^{\gamm}} \label{eq:Gammaprior_sparse}
\end{align}
where $\gamdm$ is equivalent to $\akm$, having that \eqref{eq:Wprior_sparse}, \eqref{eq:Alphaprior} and \eqref{eq:Gammaprior_sparse} form the ARD prior for each row and column of matrix $\Wm$. This way, while $\akm$ forces zero values column-wise in the latent variables, $\gamdm$ forces the zero values in the features, having a double automatic selection of relevant features.

% For labelled observations, SSHIBA can work with heterogeneous data by introducing a new observed variable binary vector $\ynm$, also of dimension $D_m$, with a conditional distribution given $\Xnm$ defined as
%  \begin{align}
%  &p\p*{\ym|\Xm} \geq h\p*{\Xm,\xim} = \nonumber\\
%  &\prodn\prodd \p*{\sigma\p*{\xindm} e^{\Xndm \yndm -\frac{\Xndm + \xindm}{2}  - \lambda \p*{\xindm} \p*{\Xndm^2 - \xindm^2}}}, \label{eq:pt}
%  \end{align}
%  where we included a lower bound to the logistic regression, $\sigma\p*{a}$ is the sigmoid function and $\lambda \p*{a} = \frac{1}{2a}\p*{\sigma\p*{a} - \frac{1}{2}}$ and $\xi^{(m)}_{n,d}$ are variational parameters that are optimised by maximising the lower bound.

% On the other hand, for the input views we will have one view for the time-independent information and the time-dependent variables will be included in different views depending on the prediction month. However, as the diagnosis is a classification problem, we included the diagnosis for previous months in views 7-11.

% \carlos{I INCLUDED THE TRANSPOSED VERSION OF THIS TABLE IN THE APPENDIX. IT IS CLEARER BUT IT FITS WORSE IN THE TEXT.}
% Therefore, we have that the first view includes the variables that are time-independent; views 2[6] the time-dependent variables corresponding to the 5 time-stamps previous to the prediction's time-stamp; views 7-11 are equivalent to the previous five but only have the diagnosis information; and views 12-14 have the three output variables, as seen in Figure \ref{fig:Scheme}.

After defining the generative model, we can evaluate the posterior distribution of all the model variables using an approximate inference approach through mean-field variational inference \cite{Blei17}. With this, we maximise a lower bound to the posterior distribution and choose a fully factorised variational family to approximate the posterior distribution as
\begin{align}
&p(\Theta|\XM) \approx  \prodn q\p*{\Zn} \nonumber \\
\eqline \prodm\p*{q\p*{\Wm} q\p*{\taum} \prodk q\p*{\akm} \prodd q\p*{\gamdm}}, \label{eq:qModel}
\end{align}
where $\Theta$ comprises all random variables in the model and $\M_i$ represents the set of views with binary data.

The mean-field posterior structure along with the lower bound 
% \carlos{\sout{in \eqref{elbo}}}
results in a feasible coordinate-ascent-like optimisation algorithm in which the optimal maximisation of  each of the factors in \eqref{eq:qModel} can be computed if the rest remain fixed using the following expression
\begin{align}\label{meanfield}
q^*(\theta_i) \propto \mathbb{E}_{\Theta_{-i}}\left[\log p(\Theta,\mathbf{x}_{1,:},\dots, \mathbf{x}_{N,:})\right],
\end{align}
where $\Theta_{-i}$ comprises all random variables but $\theta_i$. This new formulation is in general feasible since it does not require to completely marginalise $\Theta$ from the joint distribution.

\subsection{SSHIBA implementation on longitudinal data}

\begin{figure*}[thb]
  \centering
    \includegraphics[width=0.7\textwidth]{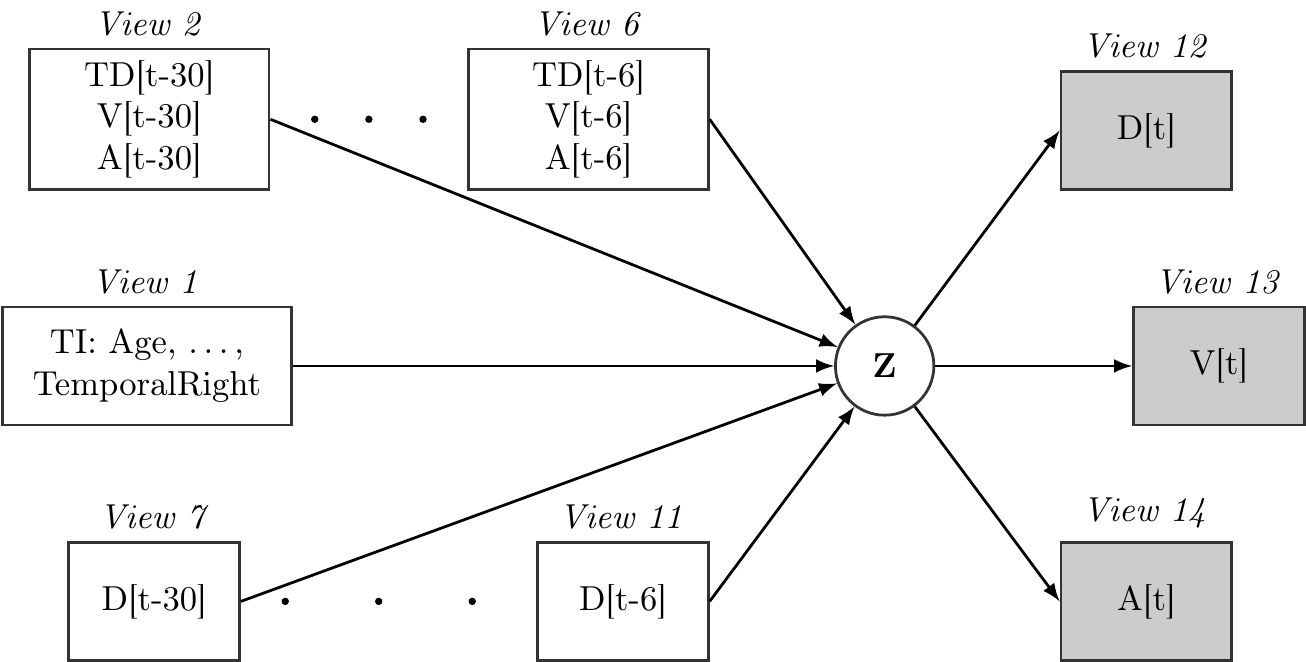}\\
  \caption{Plate diagram for the multi-view SSHIBA model adapted to longitudinal data. White circles denote unobserved random variables, white squares denote input views and grey squares output views. Note that each view is comprised by its observations and its corresponding random variables. \textit{TI} comprises the time-independent variables and \textit{TD[t]} the time-dependent variables without the \textit{Diagnosis} for month \textit{t} (\textit{D[t]}), the ventricle volume for month \textit{t} (\textit{V[t]}) and the ADAS13 score for month \textit{t} (\textit{A[t]}).[t-6], \ldots, [t-30] represent 6,\ldots, 30 months before month \textit{t}.}
  \label{fig:SchemeLongitudinal}
\end{figure*}

Taking advantage of this formulation, we propose utilising the multi-view framework to combine time-independent and time-dependent variables (as specified in Table \ref{tab:Variables}). To do so, we firstly defined one view in charge of modelling the time-independent observations, $\Xone$. Then, we combined the time-dependent data in various views based on their time-stamps. This way, we have the measures corresponding to time-stamps (6, 12, 18, 24 and 30 months before the prediction) modelled each in one view, $\{\Xtwo,\ldots,\Xsix\}$. Finally, we have one view for each prediction task $\{\Xtwelve,\Xthirteen,\Xfourteen\}$ at the desired month. Figure \ref{fig:SchemeLongitudinal} summarises the defined views where we also included views $7,\ldots,12$ to model the diagnosis as \carlos{a on-vs-all} observation. Therefore, we use the information of months 0, 6, 12, 18 and 24 to predict a result 30 months after the baseline\footnote{Note that we are doing a variable forecasting of time-stamp [t], therefore, we do not use any information of that time-stamp for the prediction task.}.

% These variables are included in three different views (output views).
This way, the model is capable of learning a projection matrix $\Wm$ for each time-stamp and a latent variable $\Z$ in charge of defining the relation between the views through the latent space. This allows the model to learn common factors corresponding to temporal information, but also to learn timestamp specific latent variables. Besides, this information can be combined with time-independent variables while analysing through this latent representation the relation with the output tasks. Therefore, in this framework, we use information of months
$0,\ldots,24$ to predict the output variables at month 30 to train the model.
For testing we use months $6,\ldots,24$ to predict the output variables at month 36, where we do not include information of month 30 to have year prediction.

\begin{table}[h!t]
% \arrayrulecolor{black} %Table line colors
% \renewcommand{\arraystretch}{1.}
\caption{Data configuration for the framework. \textit{TI} represents the time-independent data, \textit{D} represents the diagnosis, \textit{V} the ventricle volume, \textit{A} the ADAS13 score and \textit{TD} the time-dependent data for month \textit{t} including \textit{V} and \textit{A}. A hyphen implies that we are not using any information on that view for that sample and we are just using missing values. The training results corresponding to month 30 are set to missing values in order to do a one year prediction of the test set.}
\label{tab:data}
% \centering
\begin{adjustbox}{max width=\columnwidth}
\begin{tabular}{cccccccc}
\toprule
 & View & \multicolumn{6}{c}{Samples for subject \textit{n}} \\ \cmidrule(lr){3-8}
                  &                       & train 1 & train 2 & train 3 & train 4 & train 5 & test 1 \\          
\midrule
\multirow{11}{*}{Input} 
& 1 & \textit{TI} & \textit{TI} & \textit{TI} & \textit{TI} & \textit{TI} & \textit{TI}\\
& 2 & \textit{--} & -- & -- & -- & \textit{TD[0]} & \textit{TD[6]} \\
& 3 & \textit{--} & -- & -- & \textit{TD[0]} & \textit{TD[6]} & \textit{TD[12]} \\
& 4 & \textit{--} & -- & \textit{TD[0]} & \textit{TD[6]} & \textit{TD[12]} & \textit{TD[18]} \\
& 5 & \textit{--} & \textit{TD[0]} & \textit{TD[6]} & \textit{TD[12]} & \textit{TD[18]} & \textit{TD[24]} \\
& 6 &  \textit{TD[0]} & \textit{TD[6]} & \textit{TD[12]} & \textit{TD[18]} & \textit{TD[24]} & -- \\
& 7 & -- & -- & -- & -- & \textit{D[0]} & \textit{D[6]} \\
& 8 & -- & -- & -- & \textit{D[0]} & \textit{D[6]} & \textit{D[12]} \\
& 9 & -- & -- & \textit{D[0]} & \textit{D[6]} & \textit{D[12]} & \textit{D[18]} \\
& 10 & -- & \textit{D[0]} & \textit{D[6]} & \textit{D[12]} & \textit{D[18]} & \textit{D[24]} \\
& 11 & \textit{D[0]} & \textit{D[6]} & \textit{D[12]} & \textit{D[18]} & \textit{D[24]} & --\\
\midrule
\multirow{3}{*}{Output} & 12& \textit{D[6]} & \textit{D[12]} & \textit{D[18]} & \textit{D[24]} & -- & \textit{D[36]} \\
& 13 & \textit{V[6]} & \textit{V[12]} & \textit{V[18]} & \textit{V[24]} & -- & \textit{V[36]} \\
& 14 & \textit{A[6]} & \textit{A[12]} & \textit{A[18]} & \textit{A[24]} & -- & \textit{A[36]} \\
\bottomrule
\end{tabular}
\end{adjustbox}
\end{table}

In order to improve the performance of the model, we propose using the SSHIBA's missing values imputation functionality to increase the number of samples per subject used to train the model by including missing values in the months previous to the baseline. Therefore, we do not just use the information related to months 0, 6, 12, 18 and 24 to predict 30 as our training data, but also months -6, 0, 6, 12 and 18 to predict 24 and so on, where every negative month is filled with missing values. This scheme is summarised in Table \ref{tab:data}, having that the test set is always set to predict month 36 and always uses months 6 to 24 as input. \carlos{This implies that the model is calculated without using any information from month 36.} This change in the way we handle the information allows the model to have data augmentation with 5 times more training samples, $8,685$ training samples, as well as using all the available data to train the model. Table \ref{tab:data} includes the structure for the train and test set used in this article.

% We used the heterogeneity of SSHIBA to model each view differently depending on the data nature. In particular, the set of views $\{1,2,\ldots,6\}$ work with real data and also include sparsity in the features space, the set of views $\{7,\ldots,12\}$ are modelled as multi-label data and the set of views $\{13,14\}$ work with real data without sparsity in the feature space.

% SSHIBA makes automatic latent factor selection, also referred as pruning. For this purpose, d
During the inference learning we remove the $k$ columns of $\Wm$,  $\forall m$, if all the elements of $\Wkm$, across all views, are lower than the pruning threshold. For our experiments, this pruning threshold was set to $10^{-6}$. To determine the number of iterations of the inference process, we used a convergence criteria based on the evolution of the lower bound. In particular, we stop the algorithm either when $LB[-2] > LB[-1](1 - 10^{-6})$, where $LB[-1]$ is the lower bound at the last iteration and $LB[-2]$ at the previous one, or when it reaches $5\times10^{4}$ iterations. We used learning rates of $1/\#iter$, 1 and 0.9 for the projection matrices of views 7-12, 14 and the rest to adapt to the MTL problem. The SSHIBA model is available at \url{https://github.com/sevisal/SSHIBA.git}.

%%%%%%%%%%%%%%%%%%%%%%%%%%%%%%%%%%%%%%%%%%%%%%%%%%%%%%%%%%%%%%%%%%%%%%%%%%%%%%%%%%%%%%%%%%%%%%%%%%%%%%%%%%%%%%%%%%%%%%%%%%
%                                            Results 
%%%%%%%%%%%%%%%%%%%%%%%%%%%%%%%%%%%%%%%%%%%%%%%%%%%%%%%%%%%%%%%%%%%%%%%%%%%%%%%%%%%%%%%%%%%%%%%%%%%%%%%%%%%%%%%%%%%%%%%%%%

\section{Results}

This section presents the results of the prediction of clinical diagnosis, ventricle volume and the ADAS13. We additionally compare the SSHIBA based model to selected state-of-the-art methods as well as standard baseline methods. 

\subsection{Baseline methods}
We used Ridge Regression (RR) and Logistic Regression (LogR) as baseline regression and classification algorithms to analyse the imputation performance on A and V and D, respectively. 
We also selected several multitask learning based methods for comparison. First, we included a convex fused sparse group Lasso\cite{zhou2012modeling} (cFSGL) formulation. This technique encodes the temporal information by considering the sparse group Lasso penalty to select a common set of biomarkers across multiple time points and simultaneously incorporate temporal smoothness using the fused Lasso penalty. Second, we also considered multi-task techniques with concatenated temporal information:
\begin{itemize}
    \item $\ell_{1}$-norm regularized multitask learning (Least Lasso)\cite{tibshirani1996regression}: The $\ell$1-norm regularisation term captures the task relationship from multiple related tasks which introduce sparsity into the features along with the parameter for controlling the sparsity among all tasks. 
    \item Joint Feature Selection (JFS)\cite{evgeniou2007multi}: it utilises the $\ell_{1,2}$-norm regularisation term to learn sparse representations across multiple related tasks to constrain all models to share a common set of features. 
    \item Dirty Model\cite{jalali2010dirty}: it explicitly estimates a sum of two sets of parameters with multiple individual regularisation, where the corresponding matrices are encouraged to have element-wise sparsity and block-structured row sparsity.
    \item Low Rank Assumption (LRA)\cite{chen2011integrating}: it captures the task relationships using a low-rank structure, and simultaneously identifies outlier tasks using a group-sparse structure.
\end{itemize}
% Note that these methods can not work with heterogeneous data modelled in different views, and thus we included all the input views as input variables for each model and train one model for each prediction problem.
\carlos{For these methods we also considered a one-vs-all codification of the diagnosis. Therefore, they will carry out 5 task learning problem. We included all the input views as input variables for each model.} 
% can not work with heterogeneous data modelled in different views, and thus we included all the input views as input variables for each model and train one model for each prediction problem.
% The number of nearest features for both KNN and iterative imputer was set to 5.

% The main limitation for the comparative with these baselines is the fact that they can not work with heterogeneous data modelled in different views, so we had to include all the input views as input variables for each model and train one model for each prediction problem.
% The number of nearest features for both KNN and iterative imputer was set to 5.

The MALSAR package\cite{zhou2011malsar} running in MATLAB was used to implement the MTL algorithms. 
% To fine-tune the parameters, default values were used for the optional optimisation parameters such as: starting points, termination criterion, endurance, and a maximum number of repetitions. 
The regularisation parameters $\rho_{1}$ (the regularisation parameter controlling the sparsity among all tasks) and $\rho_{2}$ (an optional regularisation parameter that controls the $\ell$2-norm penalty) are selected by 10-fold cross-validation strategy on the training data. The $\rho_{1}$ and $\rho_{2}$ parameters were selected among the candidate set $\{10^{-3}, 10^{-2.5}, \ldots, 10^{2}, 2 \times 10^{2}, 2.5 \times 10^{2}, \ldots, 5 \times 10^{2}\}$ by minimising %the root-mean-square error (rmse)
the mean absolute error (MAE) 
for the regression tasks and %dice score 
balanced accuracy for the classification tasks.
The regularisation parameter $\alpha$ for RR was selected using 10-fold cross-validation from a grid of 11 values that were logarithmically spaced between -20 and 2. % INNER LOOP I HOPE. NEVERTHELESS, I WOULD POTPONE ALL IMPLEMENTATION DETAILS IN A SPECIFIC PARAGRAPH

To impute the missing data for all baselines we used 5 distinct strategies: substituting by zero, the mean, the median and the most frequent value and temporal imputation.
Temporal imputation of the missing values consists in substituting them by the mean of the subject's previous existing values if available and by the mean of the variable otherwise.

% Unlike with SSHIBA, where we assigned the variables to different views, the baselines have all the variables as input in the same matrix. 

% This project was implemented in \textit{Python 3.7} and used \textit{Scikit-learn} \cite{scikit-learn} to implement RR and LogR. 

\subsection{Metrics}

For the regression tasks, prediction of ADAS13 and Ventricle volume at month 36 (views 13 and 14), we quantified the predictive accuracy using the Mean Absolute Error (MAE) calculated as ${\rm MAE} = \frac{1}{N}\sumn\left|\mathbf{y}_{n}^{true} - \mathbf{y}_{n}^{pred}\right|$,
where $\mathbf{y}_{n}^{true}$ is the true value for sample $n$ and $\mathbf{y}_{n}^{pred}$ is the predicted value for sample $n$.
For the classification task, prediction of variable Diagnosis at month 36 (view 12), we quantified the predictive accuracy using the balanced multiclass Area Under the Curve (mAUC) \cite{zhang2020strength} calculated as ${\rm mAUC} = \frac{1}{N} \sum_c\p*{N_c \times AUC_c}$, where $N_c$ is the number of samples of class $c$ and $AUC_c$ is the AUC of class $c$ with respect to the rest of the classes.

\subsection{Performance compared to baselines}
\label{sec:aux}

In this section, we analyse the scores obtained in the prediction of the three output variables. First we compare imputation strategies in a single variable prediction problem, where we only predict variable A (The equivalent results on V and D are available in the supplementary material). Table \ref{tab:ADAS13} depicts the performance obtained by RR and SSHIBA where SSHIBA automatically imputes the missing values and RR uses distinct imputation techniques. Besides, we calculate the MAE on the prediction of ADAS13 at month 36 using various data information for the months previous to 36. Specifically, A implies that we only use ADAS13 information on previous time-stamps to predict the value at month 36, Multimodal Data (MD) represents the variables that are not ADAS13 and MD + A that we use all the available variables. 
The results determine that using temporal imputation improves the performance of RR. However, SSHIBA greatly outperforms any of the reference baselines independently of the input variables we use to train the model. Note that SSHIBA improves its performance combining all the available information whereas RR+temporal hinders it performance when using the rest of the data.

\begin{table}[thp]
\renewcommand{\arraystretch}{1.0}
\caption{Results obtained in the prediction of ADAS13 score at month 36 using information from baseline to month 24. We used MAE score as a performance measure. Columns A, MD and MD+A show the results obtained using only ADAS13 score, MD and both as input, respectively.}
\label{tab:ADAS13}
\centering
\begin{adjustbox}{max width=\columnwidth}
\begin{tabular}{cccccccc}
\toprule
\multirow{2}{*}{Regressor} & Imputation & \multicolumn{3}{c}{Input features} \\
& strategy                              & A & MD & MD + A \\
\midrule
\multirow{5}{*}{RR} 
& \textit{zero}          & 11.201 & 10.907 & 7.765\\
& \textit{mean}          & 5.766  & 5.332  & 5.194\\
& \textit{median}        & 5.957  & 5.657  & 5.234\\
& \textit{most frequent} & 8.220  & 8.095  & 6.257\\
& \textit{temporal}      & 4.045  & 4.495  & 4.258\\\midrule
\multicolumn{2}{c}{SSHIBA}           & 3.613  & 4.012  & \textbf{3.407} \\
\bottomrule
\end{tabular}
\end{adjustbox}
\end{table}

In a second experiment we compare the performance of the baselines 
% with the SSHIBA framework with a single output, where we train a model for each output variable, and 
with SSHIBA where we simultaneously predict the three output variables. Based on the previous results, we impute the baseline missing values using the temporal imputation.
\begin{table}[ht]
\renewcommand{\arraystretch}{1.}
\caption{Results of the simultaneous prediction of three output variables, A, V and D. We used two different scores for this experiment, namely, MAE for A and V and multiclass AUC for D.}
\label{tab:Results}
\centering
\begin{adjustbox}{max width=\columnwidth}
\begin{tabular}{ccccccccc}
\toprule

Model & A & V & D  \\
\midrule
Least Lasso                    & $3.623$ & $3.981$   & $0.953$  \\
JFS                            & $3.760$ &  $3.952$  &  $0.928$ \\
Dirty Model                    & $3.666$ &  $3.942$  & $0.930$  \\
LRA                            & $3.764$ & $3.984$   &  $0.933$  \\
\midrule

SSHIBA multiple output         & $\mathbf{3.406}$ & $\mathbf{2,814}$ & $\mathbf{0.956}$ \\
\bottomrule
\end{tabular}
\end{adjustbox}
\end{table}

Table \ref{tab:Results} summarises the results obtained with the analysed methods using the respective scoring to measure the performance of the models. These indicate that the proposed approach outperformed the baseline methods in the prediction of the three analysed tasks. The improvement is clearer for A and D, where the proposal outperforms the best baseline by 0.217 and 1,167, respectively.
In addition, this performance improvement was achieved while automatically imputing all the missing values in the data, having not only a prediction of the three output variables at month 36 but also for every month where there was no measure. 
% Note that the combination of the output variables in a multiple output model slightly improves the performance of SSHIBA. 

%%%%%%%%%%%%%%%%%%%%%%%%%%%%%%%%%%%%%%%%%%%%%%%%%%%%%%%%%%%%%%%%%%%%%%%%%%%%%%%%%%%%%%%%%%%%%%%%%%%%%%%%%%%%%%%%%%%%%%%%%%
%                                            Interpretability 
%%%%%%%%%%%%%%%%%%%%%%%%%%%%%%%%%%%%%%%%%%%%%%%%%%%%%%%%%%%%%%%%%%%%%%%%%%%%%%%%%%%%%%%%%%%%%%%%%%%%%%%%%%%%%%%%%%%%%%%%%%

\subsection{Analysis of the interpretability}

Factor analysis algorithms provide interpretability to the results that may help to identify relations between variables and other information related to the data. Specifically, the sparsity over the latent factors leads to having both common (similar to Canonical Correlation Analysis) and independent (similar to Principal Component Analysis) latent factors between views which, in turn, describes the relationship between variables. In this section, we analyse some of this  information learnt by the complete framework.

% Factor analysis algorithms provide interpretability to the results that may help to identify relations between variables and other information related to the data. Specifically, the sparsity over the latent factors leads to having both common and independent latent factors between views which, in turn, describes the relationship between variables. In this section we want to analyse some aspects learnt by the complete framework when it is simultaneously predicting the three variables.

\begin{figure}[h!t]
 \centering
%  \captionof{figure}{SSHIBA's kernel graphical model.}
 \includegraphics[page=1,width=\columnwidth]{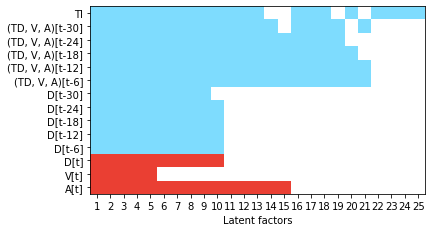}\\
 \captionof{figure}{Learnt latent factors with multiple output prediction. Coloured cells imply the latent variable is used to describe the view (blue for input data and red for output), white ones imply the latent variable is not used for this view.}
  \label{fig:latents}
\end{figure}

The ARD prior over matrix $\Wm$ induces zeros in the latent factors, leading to the elimination of some of these irrelevant factors for certain views. In particular, this pruning makes some factors to be common to certain views and not to other, therefore learning the correlation between views. Figure \ref{fig:latents} shows the latent factors learnt by the model, where the 14 views have been concatenated for the illustration and the factors have been reordered to show together those with the same active factors for the views. This image demonstrates that the model is learning 5 factors common to all views, where the information of all type of data is combined, another 5 combine all views but do not use the output ventricle information and in one case the diagnosis at the first month. Then, there are 5 views which only use the output information of the ADAS13 score and do not use the information of the diagnosis, two of which do not have the TI variables. Finally, there are 10 latent factors which do not have any output related views and combine the information of the TI and TD variables. 

Looking at the needed latent factors for each prediction task, we can see that the prediction of V is the simplest and can be done using just 5 latent factors. Equivalently, D requires 5 more latent factors for the prediction and A is the most complicated task and requires another 5 latent factors, which combine information of TI, TD, V and A.

% Comparing these results to classic feature extraction algorithms, the last 4 latent factors, which only have information of one view, could be seen as PCA (Principal Component Analysis) where the information of view 1 is used to find a low dimensional projection space. Equivalently, latent factors the other views can be seen as CCA (Canonical Correlation Analysis), where factors 16 to 21 only combine input information and the rest use some or all the views to construct a latent space using both input and output variables.

\begin{figure}[h!t]
  \centering
  \begin{subfigure}[t]{\columnwidth}
    \centering
    \includegraphics[width=0.9\columnwidth]{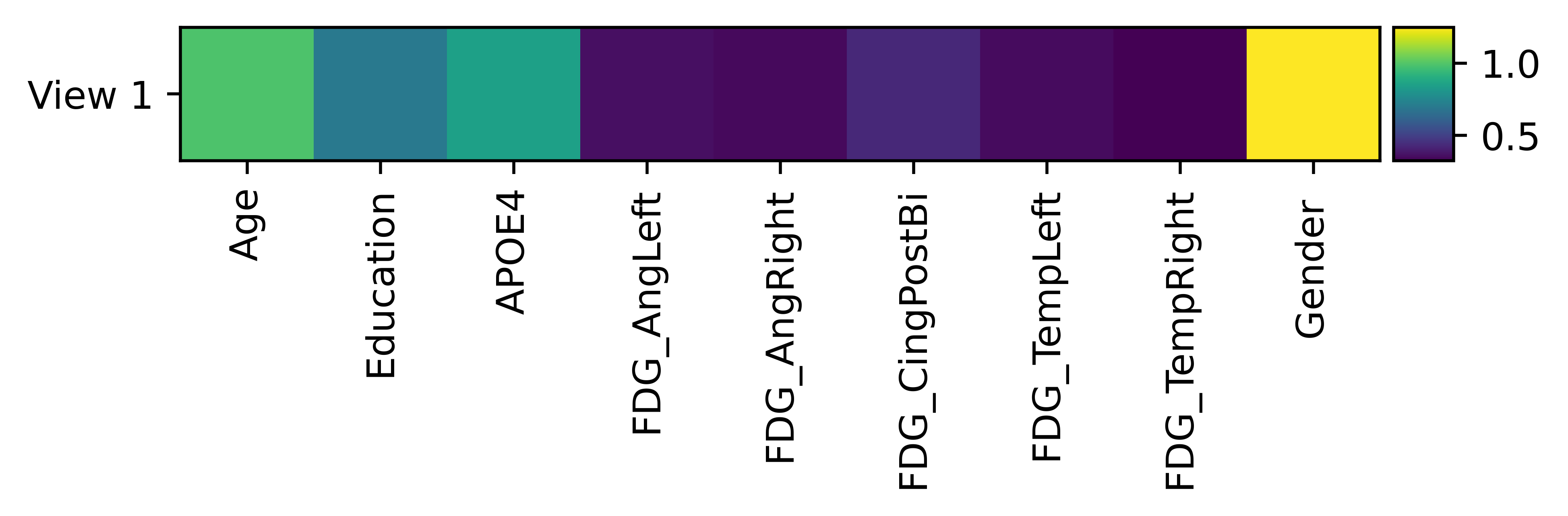}
    \caption{Relevance of TI variables.}
    \label{fig:fsTI}
  \end{subfigure}
  ~
  \begin{subfigure}[t]{\columnwidth}
    \centering
    \includegraphics[width=0.9\columnwidth]{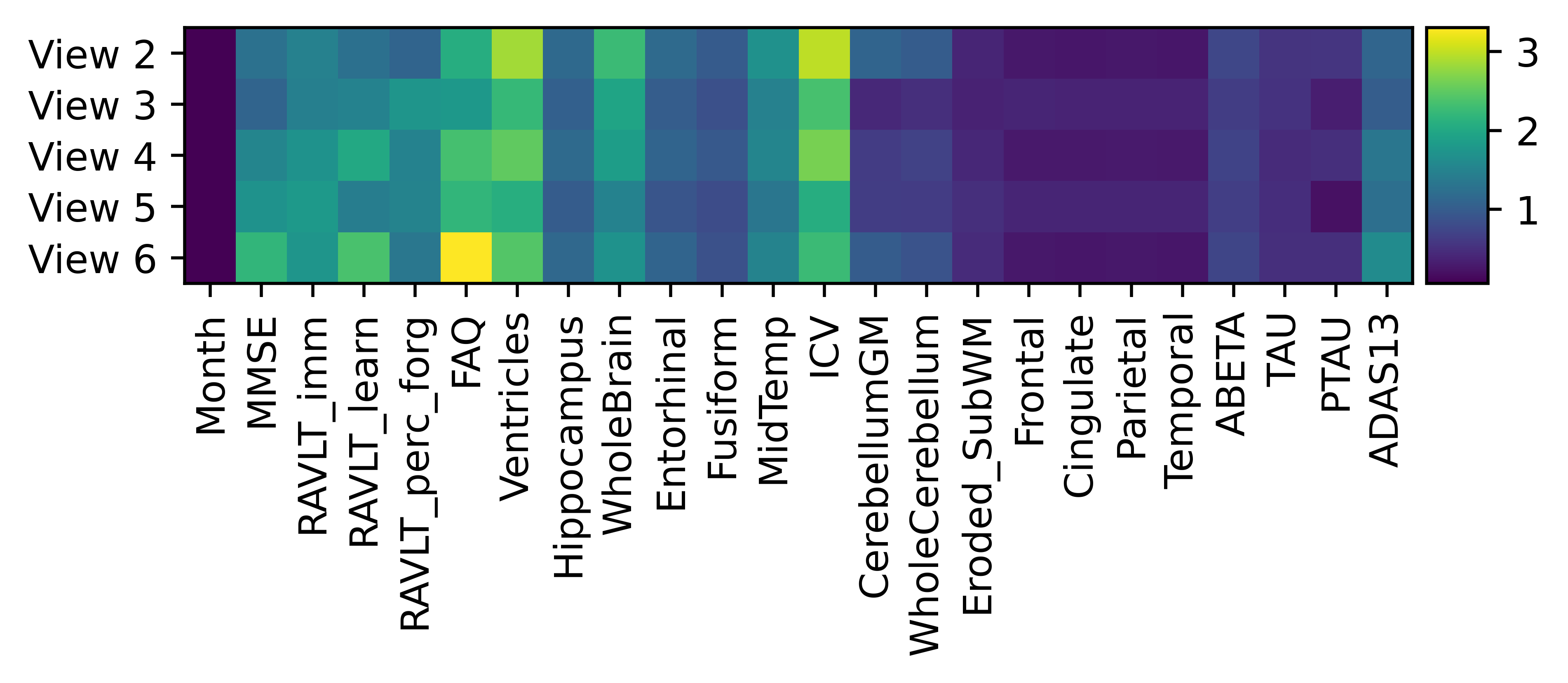}
    \caption{Month evolution of the relevance of TD variables.}
    \label{fig:fsTD}
   \end{subfigure}
  \caption{Analysis of the relevances learnt by the model for each feature. Figure \ref{fig:fsTI} represents all the time-independent variables from the model and Figure \ref{fig:fsTD} the relevance of the time-dependent data for each view.}
  \label{fig:fs2}
\end{figure}

Another functionality of the SSHIBA model is its ability to learn a relevance measure for each input variable of a certain view. The learnt relevances of the model are presented in Figure \ref{fig:fs2}, where we show the relevance learnt for the TI variables (Figure \ref{fig:fsTI}) and the TD variables for multiple time-stamps before the prediction (Figure \ref{fig:fsTD}).
These results obtain a higher relevance of the neurophysclogical and behavioural tests as well as the MRI data with respect to the rest of the variables. Furthermore, the difference in scales between both figures demonstrate that the relevance level of the TD variables make the relevance of the TI variables negligible.

However these results are highly correlated to the missing values on each data variable. Looking at the percentage of missing values in Table \ref{tab:Variables}, we can see the high number of missing values specially in the AVF45 and CSF variables, which leads to a lower feature relevance learnt by the model.
% \begin{figure}[h!t]
%     \centering
% \includegraphics[width=\columnwidth]{Images/Figure_7.png}
%   \caption{Percentage of missing values for each input variable. The values for time-dependent variables have been calculated as the mean for the analysed time-stamps of the number of missing values.}
%   \label{fig:miss2}
% \end{figure}

\begin{figure}[h!t]
  \centering
  \begin{subfigure}[t]{0.48\columnwidth}
    \centering
    \includegraphics[width=\columnwidth]{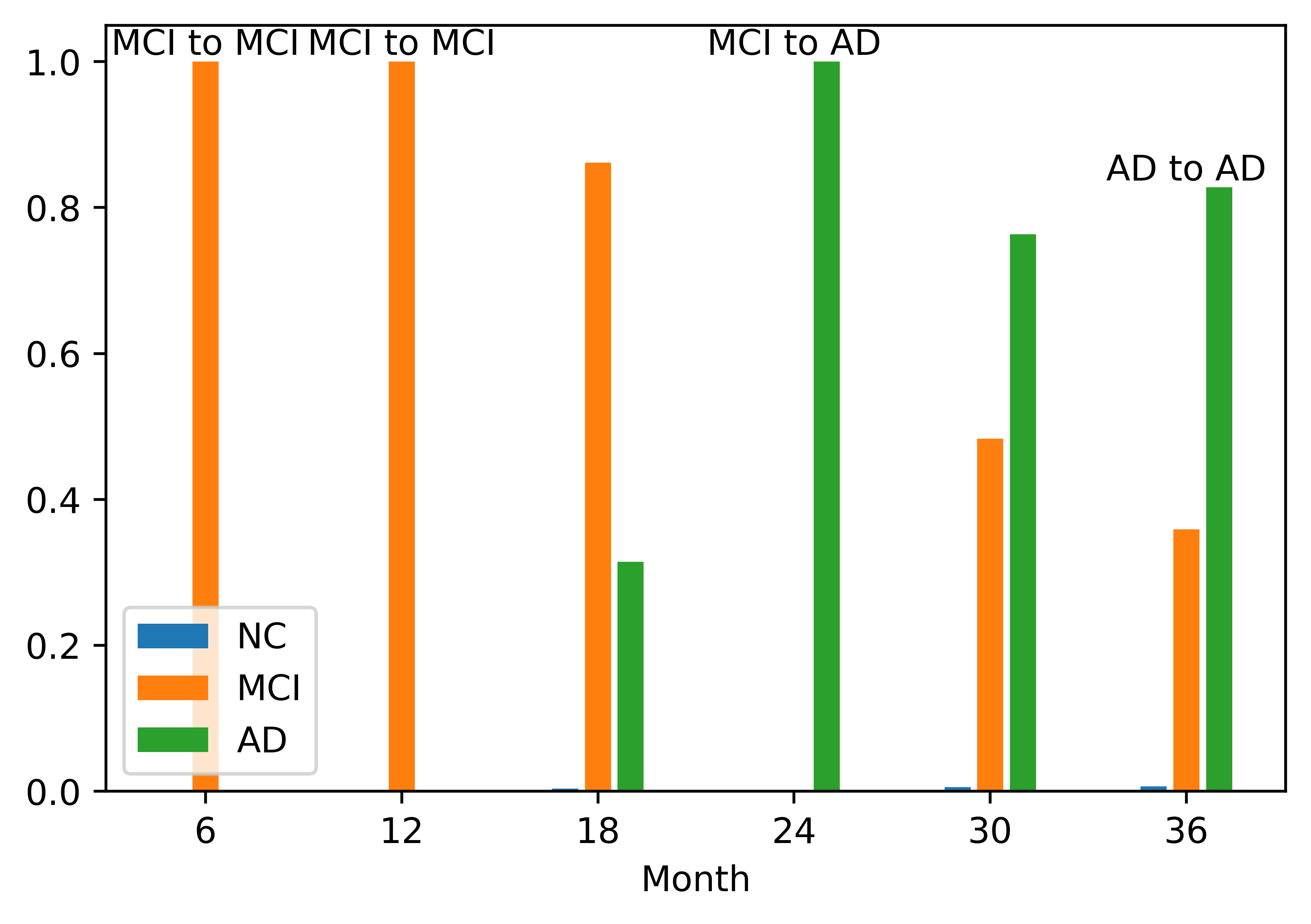}
    \caption{Prediction for subject 384}
    \label{fig:p1}
  \end{subfigure}
  ~ 
  \begin{subfigure}[t]{0.48\columnwidth}
    \centering
    \includegraphics[width=\columnwidth]{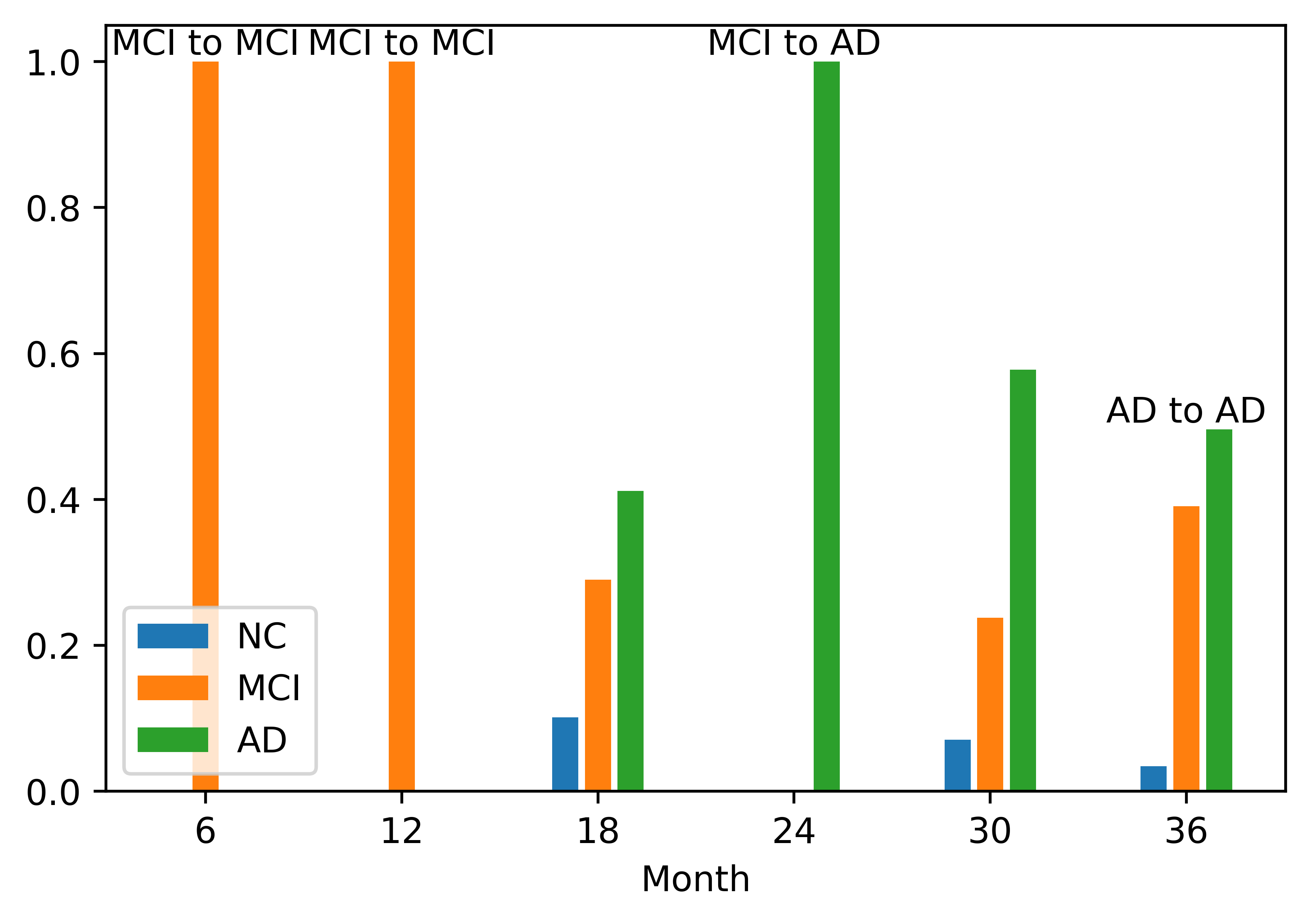}
    \caption{Prediction for subject 280}
    \label{fig:p2}
   \end{subfigure}
   \\
     \begin{subfigure}[t]{0.48\columnwidth}
    \centering
    \includegraphics[width=\columnwidth]{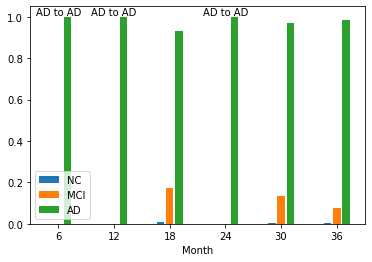}
    \caption{Prediction for subject 1281}
    \label{fig:p3}
  \end{subfigure}
  ~ 
  \begin{subfigure}[t]{0.48\columnwidth}
    \centering
    \includegraphics[width=\columnwidth]{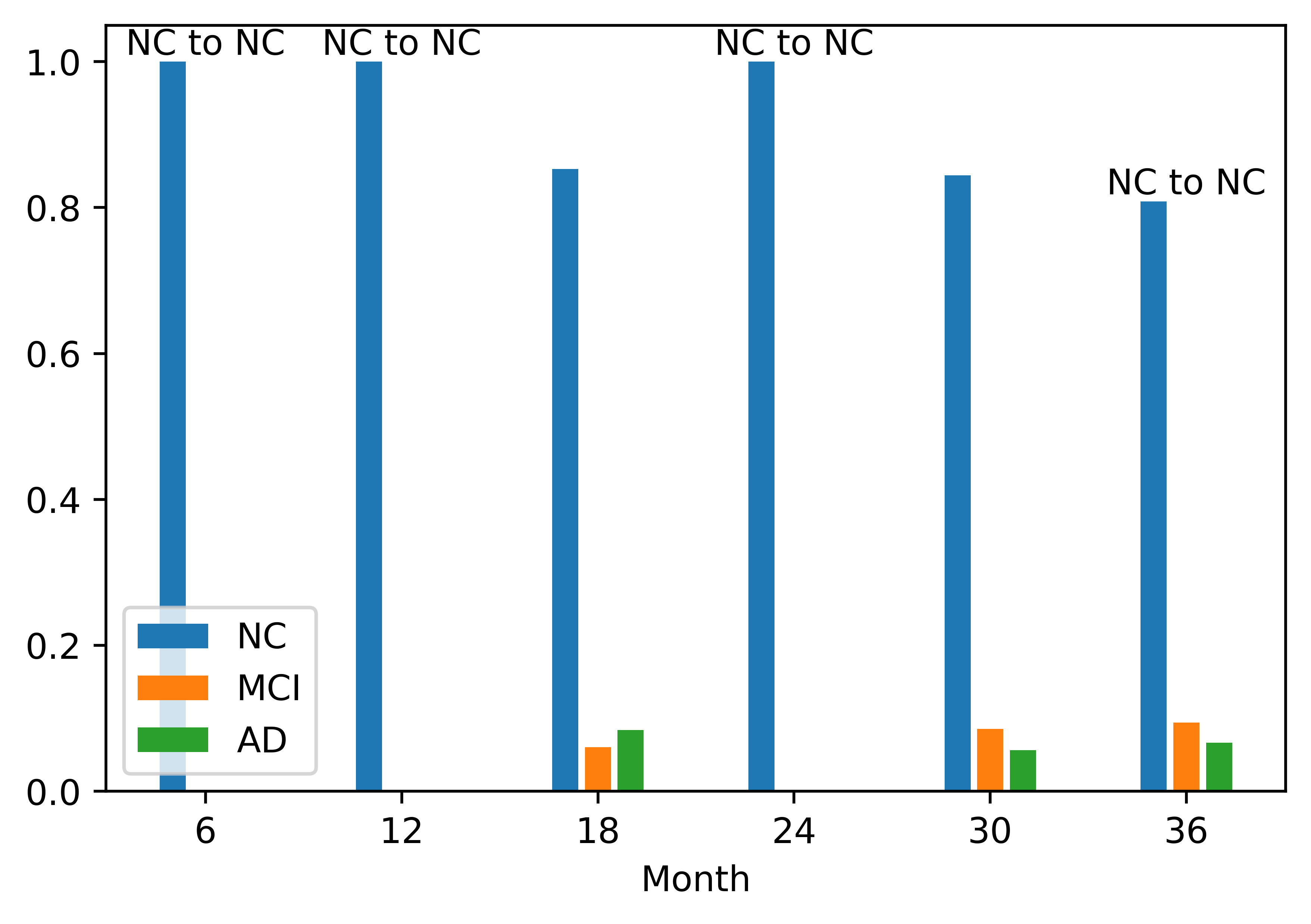}
    \caption{Prediction for subject 649}
    \label{fig:p4}
   \end{subfigure}
  \caption{Exemplary diagnosis prediction for some relevant subjects. The changes of diagnosis of the subject are shown above the bars for each month if available. If the change of diagnosis is not available it shows the predictive probability of each label learnt by the model.}
  \label{fig:subjectsDiag}
\end{figure}

Figure \ref{fig:subjectsDiag} shows some exemplary prediction results obtained with the multioutput SSHIBA framework. 
% In particular, we use the multilabel modelling of SSHIBA, making the model capable of finding two labels for a single subject in an attempt to capture the information related to the change of diagnosis of the subject. 
These images present the result of the prediction of month 36, while also showing the labels known from previous months (the ones which have a label above the bar) and the imputation of those months for which there was no subject diagnosis (no name above the bar). Looking at Figures \ref{fig:p1} and \ref{fig:p3} we see that there was a change in diagnosis between the baseline and month 36 which is also learned by the model. Specifically, we can see the label value adapts to the change of clinical diagnosis between timestamps.
% can provide a soft-diagnosis for the subject, being that it is going through a change of diagnosis.

% \begin{table}[thp]
% \renewcommand{\arraystretch}{1.2}
% \caption{Mean Absolute Error (MAE) for the ADAS13 prediction of month 36 using previous months up to month 24. In this case the data is NOT normalised.}
% \label{tab:MultiTask}
% \centering
% \begin{adjustbox}{max width=\textwidth}
% \begin{tabular}{c|c|cccccc}
% \hline
% Imputation & \multirow{2}{*}{Regressor} & \multirow{2}{*}{ADAS13} & \multirow{2}{*}{Multimodal data} & Multimodal data  \\
% strategy   &                            &                         &                           &+ ADAS13 \\\hline \hline
% \textit{zero}          & \multirow{6}{*}{Linear Reg.} & 7.970 & 9.392 & 9.797 \\
% \textit{mean}          &                              & 6.901 & 8.932 & 9.747 \\
% \textit{median}        &                              & 6.402 & 7.681 & 8.624 \\
% \textit{most frequent} &                              & 6.505 & 7.063 & 8.050 \\
% \textit{KNN}           &                              & 6.901 & 8.932 & 9.747 \\
% \textit{Iterative}     &                              & 6.901 & 8.932 & 9.747 \\
% \hline
% \hline
% \end{tabular}
% \end{adjustbox}
% \end{table}

\section{Discussion}

% Summary of the model

% model temporal information expliiting the mjultiview cofniguration, deal with missing values not only to impute missing valiarbles, but also increase the number of training data, exploit MTL learning while dealing with heterogeneous data, gain in interpretability by analyzing the latent factos, relevance of the input features and also by means of the inputation of missing diagnosis...

We analysed the viability of the application of SSHIBA for the characterisation of AD based on longitudinal data. The method uses Bayesian variational inference and allows learning the approximate posterior distribution of the model parameters to describe the observations. In this way, the algorithm is able to impute missing values in the observations, functioning in a semi-supervised manner,
% , model heterogeneous data (real and labelled data) 
and combine multiple data information in different views. In particular, it assumes that missing values in the observations correspond to random variables and then computes an approximation to their true distribution. Once the distribution has been learned, we can either sample or use a statistic, e.g., the mean, to impute unknown values. The multiview formulation allow us to combine data from individual timepoint in a single framework by using a specific projection matrix (weight matrix) for each timepoint. Therefore, the shared latent space combines the distinct data views and learns the relationships between individual timepoints. 
% This also allows to increase the number of samples used to train the model, leading to a more representative model. 
At the same time, we can use the multiview formulation to model and predict the output variables simultaneously.
% , which in our case correspond to two real variables and one categorical.

% Method novelty compared to baselines in AD
The proposed usage of SSHIBA for the characterisation of AD presents a framework that can tackle various usual problems in neuroimaging problems. Although there are baselines capable of imputing missing values, modelling longitudinal data, doing multi-task learning or learning feature relevance, SSHIBA poses a breakthrough by being able to simultaneously combining these tasks indistinctly within a single model. This, in turn, allows the model to adequately adapt to the problem needs and combine the information in a reduced latent space.

% Analysis of the results
We used SSHIBA and various baselines to model the TADPOLE database, generated with ADNI data. The results demonstrate how SSHIBA performs in imputing missing values compared to the baselines. The results indicate that SSHIBA achieves more consistent imputation, and an improvement in the prediction accuracy when all available information is combined. 
Furthermore, the comparison with state-of-the-art MTL models show a meaningful outperformance of the proposed framework in the prediction of the three tasks.

Finally, the Bayesian nature of the formulation provides further information predicting the diagnosis (NC, MCI and AD). Examining the predictions learnt by the model we see that, the algorithm captures the temporal relationships between the variables even though there is no explicitly specified temporal relation in the proposed model. This implies that the model can adapt and assign different weights to the different views (timepoints) so that their combination in the latent space describes their time dependence.

\section{Conclusion}
In this paper, we have introduced a framework based on the recently presented SSHIBA model to work in multi-tasks scenarios while work with missing values on longitudinal data. Using its multi-view capability, the model combines different time-points with various outputs in a common latent space. The results have proved that the proposed framework is adequate for high missing samples rate scenarios while greatly improving the predictive performance of the three analysed tasks with respect to the baselines. Furthermore, the results have shown that SSHIBA is able to learn the inherent variable time evolution taking advantage of the latent representation of the model.

\bibliography{bibliography}

\begin{thebibliography}{10}

\bibitem{barnes2011projected}
D.~E. Barnes and K.~Yaffe, ``The projected effect of risk factor reduction on
  alzheimer's disease prevalence,'' {\em The Lancet Neurology}, vol.~10, no.~9,
  pp.~819--828, 2011.

\bibitem{lisko2021can}
I.~Lisko, J.~Kulmala, M.~Annetorp, T.~Ngandu, F.~Mangialasche, and
  M.~Kivipelto, ``How can dementia and disability be prevented in older adults:
  where are we today and where are we going?,'' {\em Journal of internal
  medicine}, vol.~289, no.~6, pp.~807--830, 2021.

\bibitem{kueper2018alzheimer}
J.~K. Kueper, M.~Speechley, and M.~Montero-Odasso, ``The alzheimer’s disease
  assessment scale--cognitive subscale (adas-cog): modifications and
  responsiveness in pre-dementia populations. a narrative review,'' {\em
  Journal of Alzheimer's Disease}, vol.~63, no.~2, pp.~423--444, 2018.

\bibitem{sevigny2016antibody}
J.~Sevigny, P.~Chiao, T.~Bussi{\`e}re, P.~H. Weinreb, L.~Williams, M.~Maier,
  R.~Dunstan, S.~Salloway, T.~Chen, Y.~Ling, {\em et~al.}, ``The antibody
  aducanumab reduces a$\beta$ plaques in alzheimer’s disease,'' {\em Nature},
  vol.~537, no.~7618, pp.~50--56, 2016.

\bibitem{duara2008medial}
R.~Duara, D.~Loewenstein, E.~Potter, J.~Appel, M.~Greig, R.~Urs, Q.~Shen,
  A.~Raj, B.~Small, W.~Barker, {\em et~al.}, ``Medial temporal lobe atrophy on
  mri scans and the diagnosis of alzheimer disease,'' {\em Neurology}, vol.~71,
  no.~24, pp.~1986--1992, 2008.

\bibitem{burton2009medial}
E.~Burton, R.~Barber, E.~Mukaetova-Ladinska, J.~Robson, R.~Perry, E.~Jaros,
  R.~Kalaria, and J.~O’brien, ``Medial temporal lobe atrophy on mri
  differentiates alzheimer's disease from dementia with lewy bodies and
  vascular cognitive impairment: a prospective study with pathological
  verification of diagnosis,'' {\em Brain}, vol.~132, no.~1, pp.~195--203,
  2009.

\bibitem{koval2018spatiotemporal}
I.~Koval, J.-B. Schiratti, A.~Routier, M.~Bacci, O.~Colliot,
  S.~Allassonni{\`e}re, and S.~Durrleman, ``Spatiotemporal propagation of the
  cortical atrophy: Population and individual patterns,'' {\em Frontiers in
  neurology}, vol.~9, p.~235, 2018.

\bibitem{venkatraghavan2021progression}
V.~Venkatraghavan, E.~J. Vinke, E.~E. Bron, W.~J. Niessen, M.~A. Ikram,
  S.~Klein, M.~W. Vernooij, A.~D.~N. Initiative, {\em et~al.}, ``Progression
  along data-driven disease timelines is predictive of alzheimer’s disease in
  a population-based cohort,'' {\em NeuroImage}, p.~118233, 2021.

\bibitem{donohue2014estimating}
M.~C. Donohue, H.~Jacqmin-Gadda, M.~Le~Goff, R.~G. Thomas, R.~Raman, A.~C.
  Gamst, L.~A. Beckett, C.~R. Jack~Jr, M.~W. Weiner, J.-F. Dartigues, {\em
  et~al.}, ``Estimating long-term multivariate progression from short-term
  data,'' {\em Alzheimer's \& Dementia}, vol.~10, pp.~S400--S410, 2014.

\bibitem{fonteijn2012event}
H.~M. Fonteijn, M.~Modat, M.~J. Clarkson, J.~Barnes, M.~Lehmann, N.~Z. Hobbs,
  R.~I. Scahill, S.~J. Tabrizi, S.~Ourselin, N.~C. Fox, {\em et~al.}, ``An
  event-based model for disease progression and its application in familial
  alzheimer's disease and huntington's disease,'' {\em NeuroImage}, vol.~60,
  no.~3, pp.~1880--1889, 2012.

\bibitem{hardy2009missing}
S.~E. Hardy, H.~Allore, and S.~A. Studenski, ``Missing data: a special
  challenge in aging research,'' {\em Journal of the American Geriatrics
  Society}, vol.~57, no.~4, pp.~722--729, 2009.

\bibitem{atkinson2007cognitive}
H.~H. Atkinson, C.~Rosano, E.~M. Simonsick, J.~D. Williamson, C.~Davis, W.~T.
  Ambrosius, S.~R. Rapp, M.~Cesari, A.~B. Newman, T.~B. Harris, {\em et~al.},
  ``Cognitive function, gait speed decline, and comorbidities: the health,
  aging and body composition study,'' {\em The Journals of Gerontology Series
  A: Biological Sciences and Medical Sciences}, vol.~62, no.~8, pp.~844--850,
  2007.

\bibitem{marti2020survey}
G.~Mart{\'\i}-Juan, G.~Sanroma-Guell, and G.~Piella, ``A survey on machine and
  statistical learning for longitudinal analysis of neuroimaging data in
  alzheimer’s disease,'' {\em Computer methods and programs in biomedicine},
  vol.~189, p.~105348, 2020.

\bibitem{zhou2013modeling}
J.~Zhou, J.~Liu, V.~A. Narayan, J.~Ye, A.~D.~N. Initiative, {\em et~al.},
  ``Modeling disease progression via multi-task learning,'' {\em NeuroImage},
  vol.~78, pp.~233--248, 2013.

\bibitem{huang2017longitudinal}
M.~Huang, W.~Yang, Q.~Feng, and W.~Chen, ``Longitudinal measurement and
  hierarchical classification framework for the prediction of alzheimer’s
  disease,'' {\em Scientific reports}, vol.~7, no.~1, pp.~1--13, 2017.

\bibitem{adhikari2019high}
S.~Adhikari, F.~Lecci, J.~T. Becker, B.~W. Junker, L.~H. Kuller, O.~L. Lopez,
  and R.~J. Tibshirani, ``High-dimensional longitudinal classification with the
  multinomial fused lasso,'' {\em Statistics in medicine}, vol.~38, no.~12,
  pp.~2184--2205, 2019.

\bibitem{bartolucci2009examination}
A.~Bartolucci, S.~Bae, K.~Singh, and H.~R. Griffith, ``An examination of
  bayesian statistical approaches to modeling change in cognitive decline in an
  alzheimer's disease population,'' {\em Mathematics and computers in
  simulation}, vol.~80, no.~3, pp.~561--571, 2009.

\bibitem{mccombe2021practical}
N.~McCombe, S.~Liu, X.~Ding, G.~Prasad, M.~Bucholc, D.~P. Finn, S.~Todd, P.~L.
  McClean, and K.~Wong-Lin, ``Practical strategies for extreme missing data
  imputation in dementia diagnosis,'' {\em medRxiv}, pp.~2020--07, 2021.

\bibitem{zhou2012modeling}
J.~Zhou, J.~Liu, V.~A. Narayan, and J.~Ye, ``Modeling disease progression via
  fused sparse group lasso,'' in {\em Proceedings of the 18th ACM SIGKDD
  international conference on Knowledge discovery and data mining},
  pp.~1095--1103, 2012.

\bibitem{jie2015manifold}
B.~Jie, D.~Zhang, B.~Cheng, D.~Shen, and A.~D.~N. Initiative, ``Manifold
  regularized multitask feature learning for multimodality disease
  classification,'' {\em Human brain mapping}, vol.~36, no.~2, pp.~489--507,
  2015.

\bibitem{emrani2017prognosis}
S.~Emrani, A.~McGuirk, and W.~Xiao, ``Prognosis and diagnosis of parkinson's
  disease using multi-task learning,'' in {\em Proceedings of the 23rd ACM
  SIGKDD international conference on knowledge discovery and data mining},
  pp.~1457--1466, 2017.

\bibitem{imani2021comparison}
V.~Imani, M.~Prakash, M.~Zare, and J.~Tohka, ``Comparison of single and
  multitask learning for predicting cognitive decline based on mri data,'' {\em
  IEEE Access}, vol.~9, pp.~154275--154291, 2021.

\bibitem{lei2017longitudinal}
B.~Lei, F.~Jiang, S.~Chen, D.~Ni, and T.~Wang, ``Longitudinal analysis for
  disease progression via simultaneous multi-relational temporal-fused
  learning,'' {\em Frontiers in aging neuroscience}, vol.~9, p.~6, 2017.

\bibitem{cao20182}
P.~Cao, X.~Liu, J.~Yang, D.~Zhao, M.~Huang, and O.~Zaiane, ``$\ell_2$,
  1-$\ell_1$ regularized nonlinear multi-task representation learning based
  cognitive performance prediction of alzheimer’s disease,'' {\em Pattern
  Recognition}, vol.~79, pp.~195--215, 2018.

\bibitem{yang2019fused}
P.~Yang, F.~Zhou, D.~Ni, Y.~Xu, S.~Chen, T.~Wang, and B.~Lei, ``Fused sparse
  network learning for longitudinal analysis of mild cognitive impairment,''
  {\em IEEE transactions on cybernetics}, 2019.

\bibitem{tabarestani2020distributed}
S.~Tabarestani, M.~Aghili, M.~Eslami, M.~Cabrerizo, A.~Barreto, N.~Rishe, R.~E.
  Curiel, D.~Loewenstein, R.~Duara, and M.~Adjouadi, ``A distributed multitask
  multimodal approach for the prediction of alzheimer’s disease in a
  longitudinal study,'' {\em NeuroImage}, vol.~206, p.~116317, 2020.

\bibitem{sevilla2021sparse}
C.~Sevilla-Salcedo, V.~G{\'o}mez-Verdejo, and P.~M. Olmos, ``Sparse
  semi-supervised heterogeneous interbattery bayesian analysis,'' {\em Pattern
  Recognition}, vol.~120, p.~108141, 2021.

\bibitem{marinescu2018tadpole}
R.~V. Marinescu, N.~P. Oxtoby, A.~L. Young, E.~E. Bron, A.~W. Toga, M.~W.
  Weiner, F.~Barkhof, N.~C. Fox, S.~Klein, D.~C. Alexander, {\em et~al.},
  ``Tadpole challenge: Prediction of longitudinal evolution in alzheimer's
  disease,'' {\em arXiv preprint arXiv:1805.03909}, 2018.

\bibitem{prakash2020quantitative}
M.~Prakash, M.~Abdelaziz, L.~Zhang, B.~A. Strange, J.~Tohka, A.~D.~N.
  Initiative, {\em et~al.}, ``Quantitative longitudinal predictions of
  alzheimer’s disease by multi-modal predictive learning,'' {\em Journal of
  Alzheimer's Disease}, no.~Preprint, pp.~1--14, 2020.

\bibitem{duara1996alzheimer}
R.~Duara, W.~Barker, R.~Lopez-Alberola, D.~Loewenstein, L.~Grau, D.~Gilchrist,
  S.~Sevush, and P.~S. George-Hyslop, ``Alzheimer's disease: interaction of
  apolipoprotein e genotype, family history of dementia, gender, education,
  ethnicity, and age of onset,'' {\em Neurology}, vol.~46, no.~6,
  pp.~1575--1579, 1996.

\bibitem{landau2013comparing}
S.~M. Landau, M.~Lu, A.~D. Joshi, M.~Pontecorvo, M.~A. Mintun, J.~Q.
  Trojanowski, L.~M. Shaw, W.~J. Jagust, and A.~D.~N. Initiative, ``Comparing
  positron emission tomography imaging and cerebrospinal fluid measurements of
  $\beta$-amyloid,'' {\em Annals of neurology}, vol.~74, no.~6, pp.~826--836,
  2013.

\bibitem{landau2011associations}
S.~M. Landau, D.~Harvey, C.~M. Madison, R.~A. Koeppe, E.~M. Reiman, N.~L.
  Foster, M.~W. Weiner, W.~J. Jagust, A.~D.~N. Initiative, {\em et~al.},
  ``Associations between cognitive, functional, and fdg-pet measures of decline
  in ad and mci,'' {\em Neurobiology of aging}, vol.~32, no.~7, pp.~1207--1218,
  2011.

\bibitem{jagust2010alzheimer}
W.~J. Jagust, D.~Bandy, K.~Chen, N.~L. Foster, S.~M. Landau, C.~A. Mathis,
  J.~C. Price, E.~M. Reiman, D.~Skovronsky, R.~A. Koeppe, {\em et~al.}, ``The
  alzheimer's disease neuroimaging initiative positron emission tomography
  core,'' {\em Alzheimer's \& Dementia}, vol.~6, no.~3, pp.~221--229, 2010.

\bibitem{ortner2019amyloid}
M.~Ortner, R.~Drost, D.~Heddderich, O.~Goldhardt, F.~M{\"u}ller-Sarnowski,
  J.~Diehl-Schmid, H.~F{\"o}rstl, I.~Yakushev, and T.~Grimmer, ``Amyloid pet,
  fdg-pet or mri?-the power of different imaging biomarkers to detect
  progression of early alzheimer’s disease,'' {\em BMC neurology}, vol.~19,
  no.~1, pp.~1--6, 2019.

\bibitem{samuraki2007partial}
M.~Samuraki, I.~Matsunari, W.-P. Chen, K.~Yajima, D.~Yanase, A.~Fujikawa,
  N.~Takeda, S.~Nishimura, H.~Matsuda, and M.~Yamada, ``Partial volume
  effect-corrected fdg pet and grey matter volume loss in patients with mild
  alzheimer’s disease,'' {\em European journal of nuclear medicine and
  molecular imaging}, vol.~34, no.~10, pp.~1658--1669, 2007.

\bibitem{battista2017optimizing}
P.~Battista, C.~Salvatore, and I.~Castiglioni, ``Optimizing neuropsychological
  assessments for cognitive, behavioral, and functional impairment
  classification: a machine learning study,'' {\em Behavioural neurology},
  vol.~2017, 2017.

\bibitem{folstein1975practical}
M.~Folstein, ``A practical method for grading the cognitive state of patients
  for the children,'' {\em J Psychiatr res}, vol.~12, pp.~189--198, 1975.

\bibitem{rey1958examen}
A.~Rey, ``L'examen clinique en psychologie.,'' 1958.

\bibitem{pfeffer1982measurement}
R.~I. Pfeffer, T.~T. Kurosaki, C.~Harrah~Jr, J.~M. Chance, and S.~Filos,
  ``Measurement of functional activities in older adults in the community,''
  {\em Journal of gerontology}, vol.~37, no.~3, pp.~323--329, 1982.

\bibitem{johnson2013florbetapir}
K.~A. Johnson, R.~A. Sperling, C.~M. Gidicsin, J.~S. Carmasin, J.~E. Maye,
  R.~E. Coleman, E.~M. Reiman, M.~N. Sabbagh, C.~H. Sadowsky, A.~S. Fleisher,
  {\em et~al.}, ``Florbetapir (f18-av-45) pet to assess amyloid burden in
  alzheimer's disease dementia, mild cognitive impairment, and normal aging,''
  {\em Alzheimer's \& Dementia}, vol.~9, no.~5, pp.~S72--S83, 2013.

\bibitem{landau2012amyloid}
S.~M. Landau, M.~A. Mintun, A.~D. Joshi, R.~A. Koeppe, R.~C. Petersen, P.~S.
  Aisen, M.~W. Weiner, W.~J. Jagust, and A.~D.~N. Initiative, ``Amyloid
  deposition, hypometabolism, and longitudinal cognitive decline,'' {\em Annals
  of neurology}, vol.~72, no.~4, pp.~578--586, 2012.

\bibitem{shokouhi2016reference}
S.~Shokouhi, J.~W. Mckay, S.~L. Baker, H.~Kang, A.~B. Brill, H.~E. Gwirtsman,
  W.~R. Riddle, D.~O. Claassen, and B.~P. Rogers, ``Reference tissue
  normalization in longitudinal 18 f-florbetapir positron emission tomography
  of late mild cognitive impairment,'' {\em Alzheimer's research \& therapy},
  vol.~8, no.~1, pp.~1--12, 2016.

\bibitem{shaw2009cerebrospinal}
L.~M. Shaw, H.~Vanderstichele, M.~Knapik-Czajka, C.~M. Clark, P.~S. Aisen,
  R.~C. Petersen, K.~Blennow, H.~Soares, A.~Simon, P.~Lewczuk, {\em et~al.},
  ``Cerebrospinal fluid biomarker signature in alzheimer's disease neuroimaging
  initiative subjects,'' {\em Annals of neurology}, vol.~65, no.~4,
  pp.~403--413, 2009.

\bibitem{gomez2018comparison}
M.~G{\'o}mez-Sancho, J.~Tohka, V.~G{\'o}mez-Verdejo, A.~D.~N. Initiative, {\em
  et~al.}, ``Comparison of feature representations in mri-based mci-to-ad
  conversion prediction,'' {\em Magnetic resonance imaging}, vol.~50,
  pp.~84--95, 2018.

\bibitem{mohs1997development}
R.~C. Mohs, D.~Knopman, R.~C. Petersen, S.~H. Ferris, C.~Ernesto, M.~Grundman,
  M.~Sano, L.~Bieliauskas, D.~Geldmacher, C.~Clark, {\em et~al.}, ``Development
  of cognitive instruments for use in clinical trials of antidementia drugs:
  additions to the alzheimer's disease assessment scale that broaden its
  scope.,'' {\em Alzheimer disease and associated disorders}, 1997.

\bibitem{raghavan2013adas}
N.~Raghavan, M.~N. Samtani, M.~Farnum, E.~Yang, G.~Novak, M.~Grundman,
  V.~Narayan, A.~DiBernardo, A.~D.~N. Initiative, {\em et~al.}, ``The adas-cog
  revisited: novel composite scales based on adas-cog to improve efficiency in
  mci and early ad trials,'' {\em Alzheimer's \& Dementia}, vol.~9, no.~1,
  pp.~S21--S31, 2013.

\bibitem{marinescu2020alzheimer}
R.~V. Marinescu, N.~P. Oxtoby, A.~L. Young, E.~E. Bron, A.~W. Toga, M.~W.
  Weiner, F.~Barkhof, N.~C. Fox, A.~Eshaghi, T.~Toni, {\em et~al.}, ``The
  alzheimer's disease prediction of longitudinal evolution (tadpole) challenge:
  Results after 1 year follow-up,'' {\em arXiv preprint arXiv:2002.03419},
  2020.

\bibitem{neal2012Bayesian}
R.~M. Neal, {\em Bayesian learning for neural networks}, vol.~118.
\newblock Springer Science \& Business Media, 2012.

\bibitem{Blei17}
D.~M. Blei, A.~Kucukelbir, and J.~D. McAuliffe, ``Variational inference: A
  review for statisticians,'' {\em Journal of the American Statistical
  Association}, vol.~112, no.~518, pp.~859--877, 2017.

\bibitem{tibshirani1996regression}
R.~Tibshirani, ``Regression shrinkage and selection via the lasso,'' {\em
  Journal of the Royal Statistical Society: Series B (Methodological)},
  vol.~58, no.~1, pp.~267--288, 1996.

\bibitem{evgeniou2007multi}
A.~Evgeniou and M.~Pontil, ``Multi-task feature learning,'' {\em Advances in
  neural information processing systems}, vol.~19, p.~41, 2007.

\bibitem{jalali2010dirty}
A.~Jalali, S.~Sanghavi, C.~Ruan, and P.~Ravikumar, ``A dirty model for
  multi-task learning,'' {\em Advances in neural information processing
  systems}, vol.~23, pp.~964--972, 2010.

\bibitem{chen2011integrating}
J.~Chen, J.~Zhou, and J.~Ye, ``Integrating low-rank and group-sparse structures
  for robust multi-task learning,'' in {\em Proceedings of the 17th ACM SIGKDD
  international conference on Knowledge discovery and data mining}, pp.~42--50,
  2011.

\bibitem{zhou2011malsar}
J.~Zhou, J.~Chen, and J.~Ye, ``Malsar: Multi-task learning via structural
  regularization,'' {\em Arizona State University}, vol.~21, 2011.

\bibitem{zhang2020strength}
J.~Zhang, Y.~Wang, Y.~Sun, and G.~Li, ``Strength of ensemble learning in
  multiclass classification of rockburst intensity,'' {\em International
  Journal for Numerical and Analytical Methods in Geomechanics}, vol.~44,
  no.~13, pp.~1833--1853, 2020.

\end{thebibliography}

\section*{Acknowledgments}
\label{sec:acknowledgments}
This work made use of the TADPOLE data sets \url{https://tadpole.grand-challenge.org} constructed by the EuroPOND consortium \url{http://europond.eu} funded by the European Union’s Horizon 2020 research and innovation program under grant agreement No 666992.
This work was supported by Spanish MINECO (Agencia Estatal de Investigación) [RTI2018-099655-B-100 to P.O., PID2020-115363RB-I00 to C.S and V.G]; and Comunidad de Madrid [IND2017/TIC-7618, IND2018/TIC-9649, IND2020/TIC-17372, Y2018/TCS-4705 to P.O.]; and the Academy of Finland project 316258, "Predictive Brain Image Analysis", to J.T.
The work has been performed under the Project HPC-EUROPA3 (INFRAIA-2016-1-730897), with the support of the EC Research Innovation Action under the H2020 Programme to C.S.; in particular, the author gratefully acknowledges the support of A.I. Virtanen Institute for Molecular Sciences and the computer resources and technical support provided by CSC – IT Center for Science. 
Data collection and sharing for this project was funded by the Alzheimer's Disease Neuroimaging Initiative (ADNI) (National Institutes of Health Grant U01 AG024904) and DOD ADNI (Department of Defense award number W81XWH-12-2-0012). ADNI is funded by the National Institute on Aging, the National Institute of Biomedical Imaging and Bioengineering, and through generous contributions from the following: AbbVie, Alzheimer’s Association; Alzheimer’s Drug Discovery Foundation; Araclon Biotech; BioClinica, Inc.; Biogen; Bristol-Myers Squibb Company; CereSpir, Inc.; Cogstate; Eisai Inc.; Elan Pharmaceuticals, Inc.; Eli Lilly and Company; EuroImmun; F. Hoffmann-La Roche Ltd and its affiliated company Genentech, Inc.; Fujirebio; GE Healthcare; IXICO Ltd.; Janssen Alzheimer Immunotherapy Research $\&$ Development, LLC.; Johnson $\&$ Johnson Pharmaceutical Research $\&$ Development LLC.; Lumosity; Lundbeck; Merck $\&$ Co., Inc.; Meso Scale Diagnostics, LLC.; NeuroRx Research; Neurotrack Technologies; Novartis Pharmaceuticals Corporation; Pfizer Inc.; Piramal Imaging; Servier; Takeda Pharmaceutical Company; and Transition Therapeutics. The Canadian Institutes of Health Research is providing funds to support ADNI clinical sites in Canada. Private sector contributions are facilitated by the Foundation for the National Institutes of Health (\url{www.fnih.org}). The grantee organization is the Northern California Institute for Research and Education, and the study is coordinated by the Alzheimer’s Therapeutic Research Institute at the University of Southern California. ADNI data are disseminated by the Laboratory for Neuro Imaging at the University of Southern California.

\end{document}